\documentclass[10pt,twocolumn,letterpaper]{article}

\usepackage{iccv}
\usepackage{times}
\usepackage{epsfig}
\usepackage{graphicx}
\usepackage{amsmath}
\usepackage{amssymb}
\usepackage{multirow}
\usepackage{booktabs}
\usepackage{color}
\usepackage{algorithm}
\usepackage{colortbl}
\usepackage{enumitem}
\usepackage{subcaption}
\usepackage{setspace}
\usepackage[labelfont=small]{caption}
\usepackage[noend]{algpseudocode}
\usepackage{enumitem}
\usepackage[percent]{overpic}
\definecolor{customgreen}{rgb}{0.17254901960784313,0.6274509803921569,0.17254901960784313}
\definecolor{customred}{rgb}{0.8392156862745098,0.15294117647058825,0.1568627450980392}

\usepackage{authblk}

\usepackage[pagebackref=true,breaklinks=true,letterpaper=true,colorlinks,bookmarks=false]{hyperref}

\iccvfinalcopy 

\def\iccvPaperID{1082} 

\ificcvfinal\fi

\def\ie{\emph{i.e}\onedot}

\makeatletter
\def\BState{\State\hskip-\ALG@thistlm}
\makeatother

\begin{document}

\setlength\abovedisplayskip{0pt}
\setlength\belowdisplayskip{0pt}

\title{
 Preserving Modality Structure Improves Multi-Modal Learning 

}


\author{Sirnam Swetha$^1$,  Mamshad Nayeem Rizve$^1$, Nina Shvetsova$^{2,3}$, Hilde Kuehne$^{2,3,4}$, Mubarak Shah$^1$ \\%

$^1$ CRCV, University of Central Florida , 
$^2$Goethe University Frankfurt Germany, \\
$^3$University of Bonn Germany,
$^4$MIT-IBM Watson AI Lab
}


\maketitle
\ificcvfinal\thispagestyle{empty}\fi

\begin{abstract}

Self-supervised learning on large-scale multi-modal datasets allows learning semantically meaningful embeddings in a joint multi-modal representation space  without relying on human annotations. These joint embeddings enable zero-shot cross-modal tasks like retrieval and classification. However, these methods often struggle to generalize well on out-of-domain data as they ignore 
the semantic structure present in modality-specific embeddings. 
In this context, we propose a novel {Semantic-Structure-Preserving Consistency} approach to improve generalizability by preserving the modality-specific relationships in the joint embedding space. 
To capture modality-specific semantic relationships between samples, we propose to learn multiple anchors and represent the multifaceted relationship between samples with respect to their relationship with these anchors.
To assign multiple anchors to each sample, we propose a novel \textit{Multi-Assignment Sinkhorn-Knopp} 
 algorithm. Our experimentation demonstrates that our proposed approach learns semantically meaningful anchors 
in a self-supervised manner. 
Furthermore, our evaluation on MSR-VTT and YouCook2 datasets demonstrates that our proposed multi-anchor assignment based solution achieves state-of-the-art performance and generalizes to both in- and out-of-domain datasets. Code: \url{https://github.com/Swetha5/Multi\_Sinkhorn\_Knopp}

\end{abstract}


\section{Introduction}
\label{sec:intro}

\par Humans often rely on multiple sensory inputs to have a better understanding of everyday events. 
Most commonly, we utilize vision, audio, and language to perceive an event as they provide complementary information for robust reasoning. 
The closest approximation of this setup is video data as it provides both visual and audio information along with a text description as a caption. Recently, researchers have started to explore learning meaningful representations by leveraging multiple modalities to train efficient models at scale~\cite{asano2020labelling,chenbr2021multimodal, shvetsova2022everything}. 
Such systems focus on representation learning that either improves features for each modality separately~\cite{asano2020labelling} or learns a joint multi-modal embedding~\cite{chenbr2021multimodal, shvetsova2022everything} space that enables various zero-shot tasks like retrieval or classification.
However, given the inherent differences across the modalities, it is challenging to learn effective joint embeddings. Furthermore, the real-world data presents additional challenges like misalignment between modalities, leading to weak supervision. 

\par Current pre-training approaches in this area usually employ a contrastive objective~\cite{oord2018representation} to learn the joint embeddings that pulls the cross-modal embeddings of a sample from the same temporal instance closer and pushes embeddings of other samples farther. 
Despite promising performances, these methods struggle with generalizability. 
This is particularly evident in previous approaches trained on HT100M~\cite{chenbr2021multimodal, shvetsova2022everything}, which do well on the closely related downstream dataset YouCook2
but struggle to improve on the MSR-VTT dataset, which exhibits a relatively larger domain shift with respect to HT100M~\cite{shvetsova2022everything}. 
This is due to the contrastive objective's emphasis on strict alignment between modalities in the joint embedding space while ignoring the inherent weak alignment between different modalities~\cite{tang-etal-2021-decembert}, as well as the underlying semantic structure across samples~\cite{Singh_2022_CVPRflava, zellers2021merlot}. 
Recent works have tackled these issues, 
either by using joint multi-modal clustering~\cite{chenbr2021multimodal} to preserve the semantic structure in the joint embedding space or by incorporating a reconstruction objective~\cite{chenbr2021multimodal, FEEL2023aaai} to retain modality-specific features in the joint embedding space, allowing for weak multi-modal alignment. 
However, the usual reconstruction objective trivially tries to retain most modality-specific features in the joint space, thus preventing the learning of optimal features for cross-modal tasks. And the multi-modal clustering approaches perform hard-clustering making it less flexible.
Therefore, the limitations of the contrastive objective cannot be adequately addressed even after combining these independent objectives.

\par To address this, we propose a 
\underline{s}emantic-\underline{s}tructure-\underline{p}reserving \underline{c}onsistency loss {\sc (sspc)} to only retain information that is beneficial for both cross-modal embedding learning and retaining modality-specific semantic structure. In particular, 
for  {\sc sspc} loss we consider each sample (e.g., a video clip) to be composed of multiple concepts: scene or objects involved in the downstream task. 
Therefore, the relationship between samples is multifaceted, representing both shared and unique concepts across samples. 
To capture this 
multifaceted relationship in a flexible manner, we propose to learn anchors (latent codes) and model the relationship between samples with respect to their relationships with these anchors. Therefore, 
these anchors act as a proxy to represent the modality-specific relationships between samples (semantic structure) which can be preserved using the proposed {\sc sspc} loss. 
Since we have no supervision to learn these anchors, we formulate this anchor learning problem as a \emph{many-to-many} assignment problem, as modeling this multifaceted relationship simultaneously involves assigning multiple anchors to one sample and multiple samples to one anchor. Although there is a vast literature on solving the \emph{many-to-one} assignment problem~\cite{caron2020unsupervised, asano2020labelling, rizve2022towards, YM.2020Self-labelling}, there is no efficient way to solve this \emph{many-to-many} assignment problem.   

To this end,
we propose a novel \textit{Multi-Assignment Sinkhorn-Knopp} (Multi-SK) algorithm that iteratively optimizes the \emph{many-to-many} anchor assignments for both the modality-specific embeddings (in input space) and modality-agnostic multi-modal embeddings (in joint embedding space). To allow for weak alignment between modalities, we select the dominant anchors for each sample to represent the relationship between different samples.    
Our proposed {\sc sspc} loss enforces consistency between the dominant anchor assignments at the input and joint embedding spaces to preserve the modality-specific semantic structure.
To demonstrate the effectiveness of our proposed solution, we train our model on HT100M dataset and test on $6$ zero-shot tasks on multiple downstream datasets and observe that our approach leads to state-of-the-art results in all settings.

\par In summary, 
we make the following contributions: (i) We propose a flexible modality-specific semantic-structure-preserving approach to improve the generalizability of cross-modal features. (ii) We introduce \textit{ Multi-Assignment Sinkhorn-Knopp}, a novel algorithm to enable multiple  assignments for flexible sample relationship modeling. (iii) Our proposed method outperforms the current state-of-the-art for multi-modal self-supervised representation learning on \textit{both} in- and out-of domain datasets.

    

\section{Related Work}
\label{sec:rwork}

\vspace{-0.1cm}
\subsection{Multi-Modal Learning} 
\vspace{-0.1cm}

With the availability of large-scale multi-modal datasets~\cite{miech2019howto100m, sharma2018conceptual, bain2021frozen}, multi-modal learning research has received a lot of attention. It comprises of vision-language learning~\cite{vl1_radford2021learning, vl2_zhang2021vinvl}, vision-audio learning~\cite{alwassel2020selfxdc, va_1_arandjelovic2017look, va_2aytar2016soundnet, va_3chen2020vggsound, va_4_xiao2020audiovisual}, video-audio-language learning~\cite{rouditchenko2020avlnet, chenbr2021multimodal, shvetsova2022everything}, zero-shot learning~\cite{zr_huynh2020shared, zr_Mancini_2021_OpenWorld}, cross-modal generation~\cite{cmg_reed2016generative, cmg_zhou2018visual, cmg_ma2019unpaired} and multi-modal multi-task learning~\cite{mmmt_kaiser2017one}. Miech et al.~\cite{miech2019howto100m} proposed a large-scale multi-modal dataset consisting of video, audio and text by collecting instructional videos from YouTube without requiring any human annotations. Note that the text is generated from audio using Automatic Speech Recognition (ASR) and has noisy alignment between the text and video. They also proposed a multi-modal system to demonstrate the potential for learning video-text embedding via contrastive loss. To handle the noise in the dataset, Amrani et al.~\cite{amrani2021noise} proposed a noise estimation for multi-modal data via multi-modal density estimation. A noise-contrastive estimation approach in a multi-instance learning framework was 
proposed by Miech et al.~\cite{miech2020end}. XDC~\cite{alwassel2020selfxdc} performs clustering on audio-video for learning better features for each modality separately.
These works utilize only two modalities for multi-modal learning, while others have explored utilizing audio, video, and text together for multi-modal learning.  Multi-Modal versatile networks~\cite{alayrac2020selfmvn} was proposed to learn different embedding spaces for each combination of modalities. {\sc avln}et~\cite{rouditchenko2020avlnet} proposed to learn a shared embedding that maps all modalities to a single joint embedding space. Following this, {\sc mcn}~\cite{chenbr2021multimodal} proposed to perform joint clustering and reconstruction to learn joint embedding space. Note that, ~\cite{chenbr2021multimodal} performs multi-modal K-means clustering to learn hard semantic clusters. 
Unlike strict assignment in ~\cite{chenbr2021multimodal}, we propose flexible learning with multiple assignments and separately for each modality. More recently, {\sc eao}~\cite{shvetsova2022everything} utilizes transformers and combinatorial fusion of modalities to learn the joint embedding with contrastive loss. 
\par Most of these works, utilize contrastive  or clustering loss over fused multi-modal representation to learn the joint embedding space. By doing so, these models do not retain the modality-specific semantic structuring between samples encoded by the pre-trained modality specific backbones, hurting the generalization ability of the model. Additionally, recent works have reported that large-scale contrastive multi-modal models (e.g., CLIP~\cite{radford21clip}) are somewhat robust to distributional shifts mainly due to \textit{diverse} large-scale training data and prompt-engineering ~\cite{fang2022data}. Therefore, our work focuses on making the pre-training objective robust to distributional shifts. In this context, we propose a novel approach to preserve the modality-specific semantic relationships in the joint embedding space by modeling the relationship between samples w.r.t learnable anchors. To enable flexible relationship modeling between samples, we learn multiple anchor assignments per sample, where anchors shared across samples model the commonality between them, and the distinct anchors between samples highlight the uniqueness of the samples.


%


\begin{figure*}[t]
\centering
  \includegraphics[width=\linewidth]{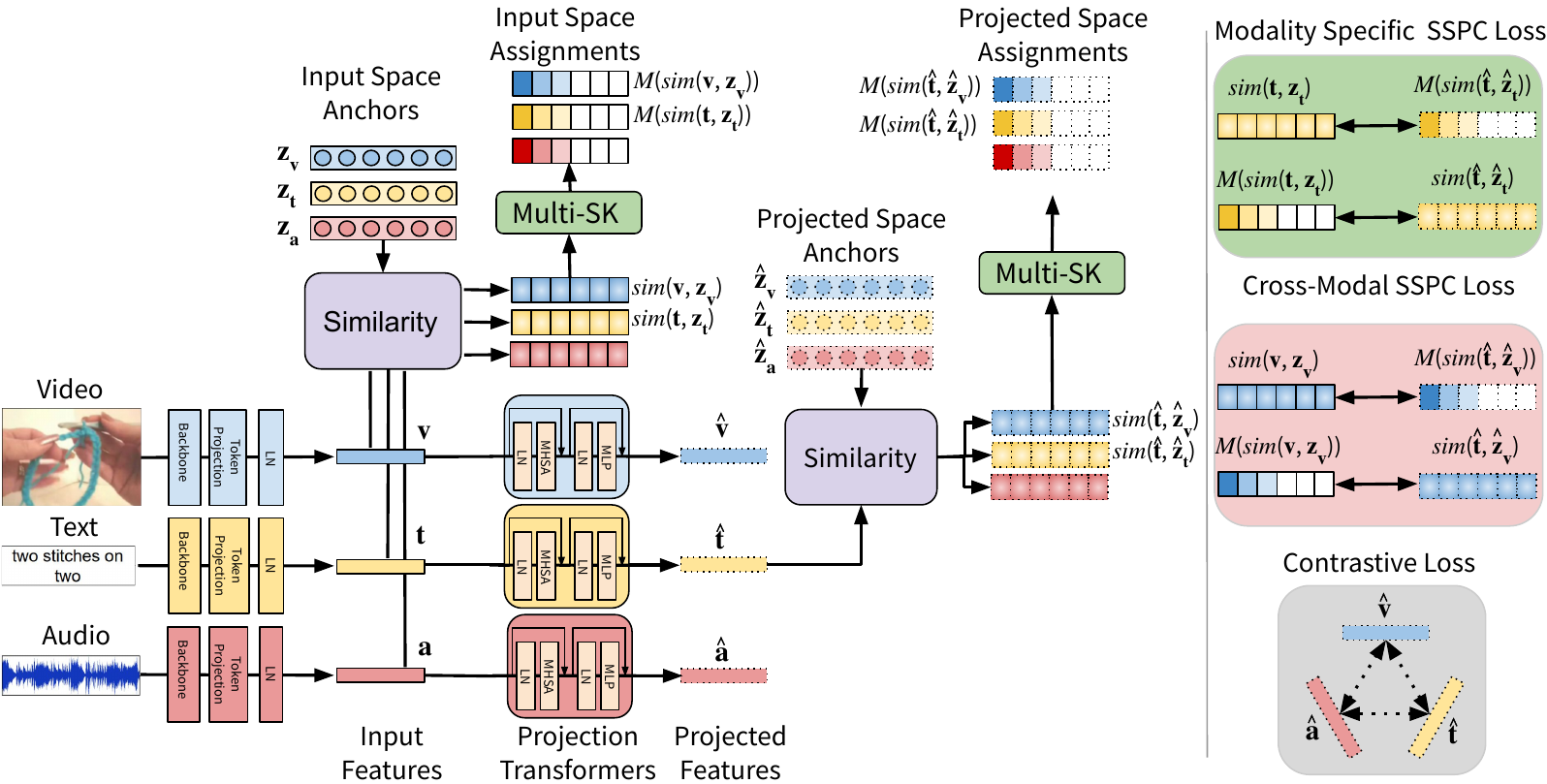}
  \vspace{-0.5em}
\caption{ \small Overview of the proposed model. Given weakly aligned text, video, and audio, we first extract features using frozen modality-specific backbones. These features are then passed through a token projection layer to obtain input features ($\mathbf{v},\mathbf{t},\mathbf{a}$) specific to each modality. Next, the modality-specific transformer models project the input features into a joint multi-modal representation space ($\mathbf{\hat{v}},\mathbf{\hat{t}},\mathbf{\hat{a}}$).
The similarity between the input features ($\mathbf{v},\mathbf{t},\mathbf{a}$) and input space anchors ($\mathbf{Z_v}, \mathbf{Z_t}, \mathbf{Z_a}$) (projected features ($\mathbf{\hat{v}},\mathbf{\hat{t}},\mathbf{\hat{a}}$) and projected space anchors($\mathbf{\hat{Z}_v}, \mathbf{\hat{Z}_t}, \mathbf{\hat{Z}_a}$)) is computed, and our Multi-SK algorithm ($M(.)$) is used to optimize multiple anchor assignments per sample, as shown in the \textit{Input Space Assignments} (\textit{Projected Space Assignments}). We do this for each modality and enforce the respective consistency losses, but in this figure we only show modality-specific consistency loss for text anchors and cross-modal consistency between text and video modalities for brevity. LN, MHSA represents LayerNorm and Multi-Head Self-Attention.
}


\vspace{-1em}
\label{fig:arch}
\end{figure*}

\subsection{Sinkhorn-Knopp}
\vspace{-0.1cm}

Recently, Sinkhorn-Knopp algorithm~\cite{sinkhorn1967concerning} has drawn huge attention because of its effectiveness in solving optimal-transport problems~\cite{kantorovich1942translation, brenier1987decomposition}. Specifically, \cite{cuturi2013sinkhorn} proposed an entropic relaxation of the optimal transport problem which can be efficiently solved using Sinkhorn's matrix scaling algorithm. Many following works have since successfully utilized the Sinkhorn-Knopp algorithm to solve different label assignment problems framed as an optimal transport problems. For instance, SeLa~\cite{YM.2020Self-labelling} 
cast the unsupervised clustering problem as a pseudo-label assignment problem and used the Sinkhorn-Knopp algorithm to solve it. SeLaVi~\cite{asano2020labelling} extended this idea to self-supervised representation learning for multi-modal data where the cluster assignments between different modalities are swapped to encourage modality invariant representation learning. 
Similarly, SwAV~\cite{caron2020unsupervised} 
used the Sinkhorn-Knopp algorithm for self-supervised representation learning and proposed to swap the pseudo-labels for differently augmented versions of a sample and use soft assignments instead of hard pseudo-labels. In contrast to these works, SuperGlue~\cite{sarlin2020superglue} used the Sinkhorn-Knopp algorithm to solve the correspondence problem between two sets of local features. Moreover, Sinkhorn-Knopp has been used in detection problems~\cite{ge2021ota}, where it was used to match the anchors with ground truths. Recently, UNO~\cite{fini2021unified}, and TRSSL~\cite{rizve2022towards} have successfully used the Sinkhorn-Knopp algorithm in solving novel class discovery and open-world semi-supervised learning problems, respectively. One key limitation of the traditional Sinkhorn-Knopp algorithm is that it cannot be directly utilized to compute multiple assignments necessary to perform multi-anchor based learning \ie \emph{many-to-many assignments}. 

Some prior works~\cite{grapa2019optimal, liu2020semantic} have attempted to solve many-to-many assignments in indirect ways.
While authors of~\cite{grapa2019optimal} use an intermediate graph to match groups of vertices from source to target graph, which can only perform group-to-group assignments and is inadequate for our problem that requires true many-to-many matching. ~\cite{liu2020semantic} modifies the Sinkhorn-Knopp row and column constraints to obtain \emph{many-to-many} assignment to model dense correspondences,  however, we have found this approach to be inferior in solving the multiple anchor assignment problem. The modified Sinkhorn-Knopp constraints approach yields suboptimal results, as discussed in Sec.~\ref{sec:ablation}. 
To address these limitations, we propose a novel algorithm Multi-SK, which outperforms the modified Sinkhorn-Knopp constraints approach to get \textit{true} many-to-many assignments.


\vspace{-0.2cm}

\section{Method}
\label{sec:method}
\par Given a set of multi-modal inputs $\{\mathbf{t}^{(i)}, \mathbf{v}^{(i)}, \mathbf{a}^{(i)}\}_{i=1}^{N}$ from N video clips, we learn modality-specific projection functions $f_t, f_v, f_a$, that transform $\mathbf{t},\mathbf{v},\mathbf{a}$ into a $d$-dimensional joint embedding space, $\mathbb{R}^d$, to obtain $\mathbf{\hat{t}}, \mathbf{\hat{v}}, \mathbf{\hat{a}}$ respectively. Our goal is to optimize the parameters of $f_t, f_v, f_a$ in such a way that they  maintain the semantic structure amongst samples from a particular modality in the joint embedding space, as discussed in Sec~\ref{sec:al}, and simultaneously brings the semantically related cross-modal inputs closer. In the following, first, we formulate our approach to modeling the relationship between samples using anchors in Sec~\ref{sec:al}, then in Sec~\ref{sec:mlsk}, we discuss our novel Multi-SK algorithm to learn these anchors for representing sample relationships, and finally, we present the overall training objective to train the model in Sec~\ref{sec:loss}.


\vspace{-0.1cm}
\subsection{Modeling Sample Relationships with Anchors}
\label{sec:al}



\par Our work aims to preserve the relationship between samples \ie $\mathbf{x}^{(i)}, \mathbf{x}^{(j)}$ from a particular modality, to better generalize on the unseen data. 
To this end, we propose to model the relationship between samples using anchors, where the similarity of each sample w.r.t. anchors encodes the semantic structure of the feature space. Unlike clustering approaches, which involve hard-assignment to a specific cluster, our approach offers flexibility \ie, there can be shared and unique anchors across samples that define the relationship between them. To be particular, for each sample, 
we learn $K$ anchors, and the similarity of assignments over these anchors represents the relationship (which we want to preserve) between samples from a particular modality.



\par  As the pre-trained features extracted from modality-specific backbones encode the semantic structure between samples within that particular modality, preserving such modality-specific relationships in the joint embedding space would boost the generalizability of the model. To achieve this, we define two sets of $K$ \textit{learnable} anchors $\mathbf{z}=\{\mathbf{z}^{(i)}\}_{i=1}^{K}$, and  $\mathbf{\hat{z}} =\{\mathbf{\hat{z}}^{(i)}\}_{i=1}^{K}$ to model the sample relationships \emph{before} and \emph{after} projecting them to the joint embedding space respectively. We repeat this for all the modalities, $t,v,a$. We propose to preserve the modality-specific semantic structure at the joint embedding space by enforcing consistency in anchor assignments before and after performing feature projections (Eq.~\ref{eq:ctl}). 

\par We still have one remaining challenge, \ie how to learn/discover these anchors in an unsupervised manner. To this end, we cast the anchor discovery as a label assignment task with a uniform prior, \ie each anchor will have an equal number of sample assignments. Additionally, to encourage flexible modeling, we enforce multiple anchor assignments per sample. However, it is difficult to estimate the exact number of anchors for each sample without additional prior information, thus we select the top $K'$ anchors for each sample to effectively model the sample relationships.
Even though this optimization task may seem like a difficult combinatorial problem, we model this combinatorial optimization task as an optimal transport problem. 
Following recent works~\cite{asano2020labelling, caron2020unsupervised}, one might assume the Sinkhorn-Knopp algorithm is a natural choice to solve this problem, however, the vanilla Sinkhorn-Knopp algorithm cannot handle multiple assignments per sample \ie many-to-many assignment. To address this limitation, we propose \textit{Multi-Assignment Sinkhorn-Knopp} algorithm, presented in the following.

\subsection{Multi-Assignment Sinkhorn-Knopp}
\label{sec:mlsk}

Given a sample matrix $\mathbf{B}$ s.t. $\mathbf{B}\in \mathbb{R}^{N\times d}$, representing $N$ samples and an anchor matrix $\mathbf{Z}$ s.t. $\mathbf{Z}\in \mathbb{R}^{K\times d}$, of $K$ anchor vectors, we obtain the similarity matrix $\mathbf{S}$ s.t. $\mathbf{S}=\mathbf{B} \mathbf{Z}^\intercal$ and $\mathbf{S}\in \mathbb{R}^{N\times K}$, where, $\mathbf{S}_{ij}$ is the probability of assigning $j$th anchor to the $i$th sample vector. The goal is to find an anchor assignment matrix, $\mathbf{Q}$, such that it satisfies the following constraints: (i) a sample should be assigned to exactly $K'$ anchors, to learn top $K'$ anchor assignments per sample. (ii) anchor assignments must be equally partitioned i.,e each anchor must be selected exactly $N\times K' / K$ times, for uniform anchor assignment.

\par To obtain such multiple anchor assignments per sample, we propose to create a substitute 3D assignment matrix $\mathbf{Q'}$ s.t. $\mathbf{Q'}\in\mathbb{R}^{K \times N \times K}$. We also generate a 3D similarity matrix $\mathbf{S'}$ from the 2D similarity matrix $\mathbf{S}$ by introducing K channels (depth dimension) to get ${K \times N \times K}$ matrix such that each channel is a scaled matrix of S with a predefined ranking between the channels enabling top $K'$ anchor selection. Since we are only interested in selecting the top $K'$ anchors, we set the first $K'$ channels of $\mathbf{S'}$ to be the same as $\mathbf{S}$. The remaining $K-K'$ channels are set to $\mu\mathbf{S}$. Here, $\mu$ is a damping factor s.t. $0  < \mu < 1$, to help select the top $K'$ anchors for each sample. 
We discuss alternate designs for $\mathbf{S'}$ generation in the Supplementary Sec. 2.


\par The optimization objective of \textit{Multi-Assignment Sinkhorn-Knopp} is to find an assignment matrix $\mathbf{Q'}$ such that it satisfies our multi-anchor assignment constraints while maximizing similarity with the initial assignment/similarity matrix, $\mathbf{S'}$. This optimization problem is defined as:
\begin{equation}
    \begin{aligned}
     \mathbf{Q^*} : \underset{\mathbf{Q'}}{\max} <\mathbf{Q'},\mathbf{S'}>~+~\epsilon H(\mathbf{Q'})\\
     H(\mathbf{Q^*}) = - \sum \mathbf{Q'}_{ijk} \log \mathbf{Q'}_{ijk}
    \end{aligned}
\end{equation}





 




\noindent $\mathbf{Q^*}$ needs to satisfy the following constraints to be a valid solution for our multi-anchor assignment problem. 

\begin{itemize}[itemsep=1pt,topsep=0pt,left=0pt]
    \item \textit{$\mathbf{Q^*}$ Row Constraint}: Within a channel, the sum of all elements in a particular row must be equal to one. This is because we only want one anchor assignment for a sample in a particular channel. 
    $\forall i,k \sum_j \mathbf{Q^*}_{ijk} = 1$
    \item \textit{$\mathbf{Q^*}$ Column Constraint}: In a channel, the sum of all elements in a column should be equal to $N/K$. This constraint enforces equal partitioning of anchor assignments.
     $\forall j,k \sum_i \mathbf{Q^*}_{ijk} = N/K$.
    
    \item \textit{$\mathbf{Q^*}$ Depth constraint}: Depth-wise sum should be equal to one for every sample and anchor combination. This constraint prevents selecting the same anchor across different channels. 
    $\forall i,j \sum_k \mathbf{Q^*}_{ijk} = 1$.
\end{itemize}

   






The traditional Sinkhorn-Knopp method uses an iterative matrix scaling algorithm that scales the rows and columns alternatively (with desired constraints) till the desired assignment matrix is obtained. We employ a similar scheme and extend the iterative scaling to the depth dimension for estimating  $\mathbf{Q^*}$. We iteratively scale the rows, columns, and channels (depth dimension) till convergence. The final 2D assignment matrix $\mathbf{Q}$ is computed by performing a depth-wise sum on the top $K'$ channels. We provide Pytorch-style pseudo-code for our Multi-Assignment Sinkhorn-Knop algorithm in the Supplementary Sec 3. 


\begin{table*}[t]
\small
    \resizebox{\textwidth}{!}{%
    \begin{tabular}{c|c|c|c|ccc|cccc|cccc}
        \toprule
    \multirow{2}{*}{Method}  
    & Retrieval & Train &
    Visual & \multicolumn{3}{c}{Trainable BB} &
    \multicolumn{4}{c}{MSR-VTT} & \multicolumn{4}{c}{YouCook2} \\ 
        \cmidrule{5-15} 
      &  & Dataset & BB & $t$ & $v$ & $a$ & {\bf R@5}$\uparrow$ & {\bf R@10}$\uparrow$ & {\bf MedR}$\downarrow$ & \textbf{MeanR}$\downarrow$ & {\bf R@5}$\uparrow$ & {\bf R@10}$\uparrow$ & {\bf MedR}$\downarrow$ & \textbf{MeanR}$\downarrow$ \\ 
    
    \midrule
    
    {\sc A}ctBERT~\cite{zhu2020actbert} &   t $\rightarrow$ v &  HT100M &  Res3D+Faster R-CNN & & &  & 23.4 & 33.1 & 36 & - & 26.7 & 38.0  & 19 & - \\
    SupportSet~\cite{patrick2020support} &   t $\rightarrow$ v &  HT100M &  R152 + R(2+1)D-34 & \checkmark & &  & 23.0 & 31.1 & 31 & - & - & -  & - & - \\
    HT100M~\cite{miech2019howto100m} &   t $\rightarrow$ v &  HT100M &  {\sc R152 + RX101} &  & &  & 21.2 & 29.6 & 38 & - & 17.3 & 24.8  & 46 & - \\
    {\sc avln}et~\cite{rouditchenko2020avlnet} &   t $\rightarrow$ v &  HT100M & {\sc R152 + RX101} & & & \checkmark & 24.7 & 34.2 & - & - & 21.1 & 29.6  & - & - \\
    {\sc eao}~\cite{shvetsova2022everything} &   t $\rightarrow$ v &  HT100M & {\sc R152 + RX101} & & & \checkmark & 24.6 & \textbf{35.3} & 25 &\textbf{ 90.4} & \underline{27.9} & \underline{38.9}  & \underline{19} & \underline{119.6} \\
     
    \rowcolor[gray]{.95} {Ours} &   t $\rightarrow$ v &  HT100M & {\sc R152 + RX101} & & & \checkmark &  \textbf{26.4} & \underline{35.1} & \textbf{23} & \underline{92.2} & \textbf{29.4} & \textbf{40.7}  & \textbf{18} & \textbf{111.8}\\
    \midrule
    {\sc AVLN}et~\cite{rouditchenko2020avlnet} &   v $\rightarrow$ t  &  HT100M & {\sc R152 + RX101} & & & \checkmark & \underline{27.2} & 35.7 & \underline{25} & 86.5 & 22.8 & 32.9  &  30 & 142.2\\
    {\sc eao}~\cite{shvetsova2022everything} &   v $\rightarrow$ t &  HT100M & {\sc R152 + RX101} & & & \checkmark & \textbf{27.6} & \underline{36.6} &  \underline{25}  & \underline{85} & \underline{31.8} & \underline{70.5} & 15 & \underline{91.9}\\
    
    \rowcolor[gray]{.95} {Ours} &   v $\rightarrow$ t &  HT100M & {\sc R152 + RX101} & & & \checkmark & \underline{27.2} & \textbf{37.1} & \textbf{23} & \textbf{84.5} & \textbf{32} & \textbf{72} & 15 & \textbf{85.2} \\
    
    \midrule
    {\sc AVLN}et~\cite{rouditchenko2020avlnet} & t $\rightarrow$ v + a &  HT100M & {\sc R152 + RX101} & & & \checkmark & 19.2 & 27.4 & 47 & - &  36.1 & 44.3 & 16 & -  \\
    {\sc MCN}~\cite{chenbr2021multimodal} &   t $\rightarrow$ v + a &  HT100M & {\sc R152 + RX101} & & & \checkmark & \textbf{25.2} & \underline{33.8} & - & - & 35.5 & 45.2  & -  & - \\
    {\sc eao}~\cite{shvetsova2022everything} &   t $\rightarrow$ v + a &  HT100M & {\sc R152 + RX101} & & & \checkmark & 23.3  & 33.2  & \underline{29} & \underline{94.8} & \underline{38.5} & \underline{49.2} & \underline{11}  & \textbf{82.7}\\
     
    \rowcolor[gray]{.95} {Ours} &   t $\rightarrow$ v + a &  HT100M & {\sc R152 + RX101} & & & \checkmark & \underline{25.1} & \textbf{34.5} & \textbf{26} & \textbf{91.8} & \textbf{39.4}  & \textbf{50.1} & \textbf{10} & \underline{83.3} \\
    \midrule
    
    {\sc AVLN}et~\cite{rouditchenko2020avlnet} &  v + a $\rightarrow$ t &  HT100M & {\sc R152 + RX101} & & & \checkmark & 19 & 26.3 & 44 & 128.1 & \underline{48.8} & 58.4 & 6  & 67.1 \\
    {\sc eao}~\cite{shvetsova2022everything} &   v + a $\rightarrow$ t &  HT100M & {\sc R152 + RX101} & & & \checkmark & \underline{21.8}  & \underline{31.4}  & \underline{28.5} & \underline{98.9} & \textbf{49} & \underline{60.9} & 6  & \underline{43.8}\\
     
   \rowcolor[gray]{.95} {Ours} &   v + a $\rightarrow$ t &  HT100M & {\sc R152 + RX101} & & & \checkmark & \textbf{24} & \textbf{32} & \textbf{27} & \textbf{95.9} & \underline{48.8}  & \textbf{61.3} & 6 & \textbf{43.5} \\

    \bottomrule
    \end{tabular}%
}
    \vspace{-0.5em}
     \caption{{\small Zero-shot Retrieval results on MSR-VTT/YouCook2. For fair comparison, we compare with models trained on 
     text, video and audio. 
     Retrieval column represents the evaluation task. BB=Backbone. \textbf{Bold}, \underline{underline} represent highest and second-highest scores.}}
    \label{tab:zs_ret_tva}
    \vspace{-1em}
\end{table*}

\subsection{Training Objective}
\label{sec:loss}
\noindent \textbf{Semantic Structure Preserving Consistency Loss.} To preserve the semantic structure of each modality, $t,v,a$, we apply consistency loss to enforce similar anchor assignments between input and joint embedding space. As the cross-modal contrastive loss in the joint embedding space tries to bring different modalities together, features in the joint embedding space from a particular modality should preserve the common anchors that exist in corresponding features from the other modalities. Therefore, we also apply cross-modal anchor consistency across all modalities as shown in Eq.~\ref{eq:ctl}. Since we are dealing with $3$ input modalities, this cross-modal consistency  results in $9$ consistency constraints.


Let's denote $\mathcal{L}(\mathbf{t}, \mathbf{\hat{v}}, \mathbf{z}_t, \mathbf{\hat{z}}_t)$ as the consistency loss between text anchor assignment in the input space and the textual anchor assignment of the corresponding video features at the joint embedding space as shown  below in Eq.~\ref{eq:ctl_tvt}:



\vspace{-0.5em}
\begin{equation}
     \label{eq:ctl_tvt}
  \begin{aligned}
  \mathcal{L}(\mathbf{t}, \mathbf{\hat{v}}, \mathbf{z}_t, \mathbf{\hat{z}}_t) = \alpha_{\mathbf{t}, \mathbf{\hat{v}}} {g}(\mathrm{sim}(\mathbf{t}, \mathbf{z}_t), M(\mathrm{sim}(\mathbf{\hat{v}}, \mathbf{\hat{z}}_t)) ) \\
  + \beta_{\mathbf{t}, \mathbf{\hat{v}}} {g}(\mathrm{sim}(\mathbf{\hat{v}}, \mathbf{\hat{z}}_t)), M(\mathrm{sim}(\mathbf{t}, \mathbf{z}_t))), 
  \end{aligned}
\end{equation}
Here, $\mathbf{z}_t, \mathbf{\hat{z}}_t$ respectively represent input and output learnable anchor  vectors for the text modality. $g(.)$, and $M(.)$ respectively represent binary cross-entropy-with-logits loss  and Multi-Assignment Sinkhorn-Knopp (discussed in Sec.~\ref{sec:mlsk}), $\alpha$ and $\beta$ represents loss coefficients, and $\mathrm{sim}(\mathbf{a}, \mathbf{b})=\exp(\mathbf{a}.\mathbf{b}/\tau\lVert \mathbf{a} \rVert\lVert \mathbf{b} \rVert)$, $\tau$ is the temperature hyperparameter of the similarity metric.

\noindent Overall semantic structure preserving consistency loss for all the modalities is defined as:
\vspace{-0.5em}



\begin{equation}
    \label{eq:ctl}
    \begin{aligned}
     \mathcal{L}_{sspc} = \sum_{m\in\{t,v,a\}} \sum_{n\in\{t,v,a\}}\mathcal{L}(\mathbf{m}, \mathbf{\hat{n}}, \mathbf{z}_m, \mathbf{\hat{z}}_m).
    \end{aligned}
\end{equation}

\noindent \textbf{Contrastive Loss.} Following~\cite{chenbr2021multimodal, shvetsova2022everything}, we also use contrastive loss to bring cross-modal embeddings of the same sample closer while pushing away embeddings from other sample. For this, we use $3$ pairwise single-modality constrastive losses, $\mathcal{L}_{nce\_tv}, \mathcal{L}_{nce\_ta}, \mathcal{L}_{nce\_va}$ between $(t,v), (t,a), (v,a)$ respectively. Specifically, we use Noise Contrastive Estimation~\cite{oord2018representation} with temperature $\kappa$ as shown in Eq.~\ref{eq:cl_nce}. 


\begin{equation}
 \label{eq:cl_nce}
\begin{aligned}
    \mathcal{L}_{nce\_xy}= - \log \cfrac{\exp(\mathbf{x}^\intercal \mathbf{y}/\kappa)}{\sum_{i=1}^{N} \exp(\mathbf{x}^{(i)\intercal} \mathbf{y}^{(i)}/\kappa)}
\end{aligned}    
\end{equation}

The overall contrastive loss for all modalities is defined as $\mathcal{L}_{nce} = \lambda_{tv}\mathcal{L}_{nce\_tv} + \lambda_{ta} \mathcal{L}_{nce\_ta} +  \lambda_{va}\mathcal{L}_{nce\_va}$
   


\section*{Overall Loss} The overall training objective is a combination of {\sc sspc} loss (Eq.~\ref{eq:ctl}) and contrastive loss (\ref {eq:cl_nce}):$\mathcal{L}_f = \lambda_{sspc} \mathcal{L}_{sspc}+\lambda_{nce}\mathcal{L}_{nce}$, where, $\lambda_{sspc}$ and $\lambda_{nce}$ are loss coefficients. By combining both losses, the model learns a more generic joint embedding space which preserves the modality-specific semantic structure by enforcing the anchor assignment similarity before and after feature projection and  also brings representations of different modalities together by utilizing contrastive loss. 

\section{Experiments}
\label{sec:exp}

\subsection{Experimental Setup}
\label{sec:exp_setup}
\noindent{\bf Backbones.} 
For comparability, we follow the same setup of previous works~\cite{miech2019howto100m, rouditchenko2020avlnet, chenbr2021multimodal, shvetsova2022everything}. As visual backbone, we use a combination of 2D features from ResNet-152~\cite{he2016resnet} pretrained on Imagenet~\cite{deng2009imagenet}, and 3D features from ResNeXt101~\cite{hara2018canresnext} pretrained on Kinetics~\cite{carreira2017kinetics}. The text backbone is  GoogleNews pretrained Word2vec model~\cite{mikolov2013efficient}. These backbones are fixed and not finetuned during training. Following~\cite{chenbr2021multimodal, shvetsova2022everything}, we use a trainable CNN with residual layers as an audio backbone. We provide additional details in the Supplementary Sec. 5. 

\noindent{\bf Data Sampling.} We use a batch of 216 videos and randomly sample ten 8-second clips per video. If the sampled clip contains narration (95\% clips), we use ASR time stamps to select clip borders. To disentangle high text-audio
correlation in HT100M, 
we shift the audio clip randomly by 4 seconds with respect to the video and text boundaries.

\noindent{\bf Projections.} Following~\cite{shvetsova2022everything, chenbr2021multimodal,rouditchenko2020avlnet}, we use a gated linear projection~\cite{miech2017gated} to project features into common token space, as well as to project resulting tokens into shared embedding space. We set the dimension of the common token space to 4096 and of the shared embedding space to 6144. We use a single transformer block with hidden size of 4096 with 64 heads and an MLP size of 4096. We set the number of anchors $K$ as $64$ and  $K'$ as $32$ with damping factor $\mu$ as $0.25$. We train all models for 15 epochs using an Adam optimizer~\cite{kingma2014adam} with a learning rate of 5e-5, exponential decay of 0.9 and the temperature of cosine similarity $(\tau)$ as 0.1. We maintain a memory-bank of size $5500$ while performing Multi-SK. Following~\cite{alayrac2020selfmvn,shvetsova2022everything}, we set higher weight for loss terms that involve text-video in Eq.~\ref{eq:ctl}, \ie, for all text-video terms the weight is set to 1.0 and the rest of the weights are set to 0.1. 

\subsection{Datasets, Tasks, Metrics}
\label{sec:exp_data}

\noindent{\bf Pretraining Dataset.} We train our model on the HT100M dataset~\cite{miech2019howto100m}, which contains over 1 million instructional videos with automatically generated text narrations. The text narrations can be assumed to be noisy and to not always describe the video scene.

\begin{table*}[t]
\small
    \resizebox{\textwidth}{!}{%
    \begin{tabular}{c|c|c|c|ccc|cccc|cccc}
        \toprule
    \multirow{2}{*}{Method}  
    & Retrieval & Train &
    Visual & \multicolumn{3}{c}{Trainable BB} &
    \multicolumn{4}{c}{MSR-VTT} & \multicolumn{4}{c}{YouCook2} \\ 
        \cmidrule{5-15} 
      &  & Dataset & BB & $t$ & $v$ & $a$ & {\bf R@5}$\uparrow$ & {\bf R@10}$\uparrow$ & {\bf MedR}$\downarrow$ &  {\bf MeanR}$\downarrow$ & {\bf R@5}$\uparrow$ & {\bf R@10}$\uparrow$ & {\bf MedR}$\downarrow$ & {\bf MeanR}$\downarrow$ \\ 
    
    \midrule
    
    {\sc avln}et~\cite{rouditchenko2020avlnet} &  t $\rightarrow$ v &  HT100M & {\sc R152 + RX101} &  & & \checkmark & 42.2 &  56.2 & 7 & \underline{35.1}  & 23.7 & 32.7 & 28  & 122.6  \\
    {\sc eao}~\cite{shvetsova2022everything} &    t $\rightarrow$ v &  HT100M & {\sc R152 + RX101} & & & \checkmark &  \underline{47.7}  & \underline{59.3} & \underline{6} & 35.6 & \underline{33.9} & \underline{45.8} & \underline{13} & \underline{70.7}  \\
   \rowcolor[gray]{.95} {Ours} & t $\rightarrow$ v &  HT100M & {\sc R152 + RX101} & & & \checkmark & \textbf{48.7} & \textbf{60.6} & \textbf{5} & \textbf{33.1} & \textbf{35.6} &  \textbf{48.1} & \textbf{12} & \textbf{64.9} \\
        \bottomrule
    \end{tabular}}
    \vspace{-0.5em}
    \caption{{\small Text-to-video retrieval results on MSR-VTT/YouCook2 in the fine-tune setting. For fair comparison, we compare with  models trained on all 3 modalities \ie text, video and audio. \textbf{Bold}, \underline{underline} represent best and second-best scores.}}
    \label{tab:ft_ret}
    \vspace{-1em}
\end{table*}
\noindent{\bf Zero-shot Retrieval.} We use MSR-VTT~\cite{xu2016msrvtt} and YouCook2~\cite{zhou2018youcook} datasets to evaluate the zero-shot retrieval capability of our model. We report performance on $4$ retrieval tasks: (i) Text-to-Video retrieval, (ii) Video-to-Text retrieval, (iii) Text-to-Video-Audio retrieval, (iv) Video-Audio-to-Text retrieval. The YouCook2 dataset contains cooking videos from YouTube with human-annotated clips ($ \sim 2$ - $200$ secs). For evaluation we use at maximum first 48 seconds of clip, since most clips are shorter than that. The MSR-VTT dataset contains human-annotated clips ($ \sim 10$ - $30$ secs) on various topics and provides captions with natural language sentences. Following ~\cite{miech2019howto100m,rouditchenko2020avlnet,chenbr2021multimodal, shvetsova2022everything}, to evaluate our model on MSR-VTT, we use the 1k set of test clips~\cite{yu2018jointfusion}, and for YouCook2, we use 3,350 validation clips~\cite{miech2019howto100m}. To perform $(t \rightarrow v+a)$ retrieval, we compute similarities by dot product between a text query t and all videos in the dataset using a averaged $v+a$ representation for each video. We report standard recall metrics R@5, R@10, median rank (MedR) and the mean rank(MeanR). Further, we also evaluate our model using CLIP backbone and report results in Sec. 1 of the Supplementary.

\noindent{\bf Zero-Shot Full-Video Retrieval.} Following ~\cite{chenbr2021multimodal}, we evaluate Zero-Shot Full Video Retrieval from a set of captions on YouCook2 dataset. We report recall metrics following Caption averaging method ~\cite{chenbr2021multimodal}that finds maximal prediction over all the clips of video for each caption and averaging over set of captions in query leading to a single prediction for full video.


\noindent{\bf Text-to-video Retrieval after Fine-tuning.} We additionally evaluate the retrieval performance of the models finetuned on downstream tasks. We use $6783$ clips from MSR-VTT (which contain audio) and $9586$ clips from YouCook2 train datasets to fine-tune the model as proposed by~\cite{rouditchenko2020avlnet}.

We also evaluate our model for Zero-shot Classification 
and report additional results in Sec. 1 of the Supplementary.

\subsection{Comparison with State-of-the-art}
\label{sec:exp_sota}

\noindent{\bf Zero-shot Retrieval Tasks.} In Tab.~\ref{tab:zs_ret_tva}, we report the performance of the learned multi-modal representations on four zero-shot retrieval tasks, (i) Text-to-Video, (ii) Video-to-Text, (iii) Text-to-Video-Audio, (iv) Video-Audio-to-Text., on MSR-VTT and YouCook2 datasets. For a fair comparison, we only compare with models trained on all three modalities \ie text, video and audio. In summary, our proposed method outperforms the current state-of-the-art methods by a noticeable margin on \textit{both} the datasets. The results on the MSR-VTT dataset are particularly interesting since it demonstrates the generalizability of our model. In particular, HT100M  consists of instructional videos and the textual descriptions are generated from audio using ASR. The text narration has noisy alignment with video and typically describes the steps in the video.
YouCook2 shares domain similarity with HT100M as its videos are instructional format, and the text descriptions corresponds to specific steps in the recipe. In contrast, MSR-VTT is not restricted to instructional videos and the text is typically a single sentence caption describing the whole video. As a result, there is a distributional shift between HT100M and MSR-VTT. Therefore, zero-shot retrieval on MSR-VTT has to overcome distributional shift, which is a crucial requirement for practical deployment. Our proposed method shows relatively higher improvement on this task validating the effectiveness of anchor-based learning.

In case of text-to-video-audio retrieval, our method improves the Median and Mean rank of the baseline~\cite{shvetsova2022everything} by $3\%$ on MSR-VTT along with gains on recall metrics $R@5$ and $R@10$. For video-audio-to-text retrieval, our methods performs very well on MSR-VTT with $3\%$ improvement on MeanR, $2.2\%$ improvement on $R@5$. We also observe similar improvements on the YouCook2 dataset. Similarly, our proposed method outperforms the current state-of-the-art on most of the metrics for text-video-retrieval video-text-retrieval.
Additionally, to compare our approach with \textit{text-video} only model, we train our model on video and text to compare the performance with state-of-the-art in Sec 1 of the Supplementary.

    

\begin{table}
\small
    \resizebox{\columnwidth}{!}{\begin{tabular}{ccccc}
        \toprule
    Method & Aggregation
    & {\bf R@1}$\uparrow$ & {\bf R@5}$\uparrow$ & {\bf R@10}$\uparrow$   \\
    \midrule
    Random & - & 0.23 & 1.15 & 2.32 \\
    HT100M ~\cite{miech2019howto100m} & Caption Avg. & 43.1& 68.6& 79.1 \\
    {\sc mil-nce} ~\cite{miech2020end} & Caption Avg. & 46.6& 74.3& 83.7 \\
    {\sc mcn} ~\cite{chenbr2021multimodal} & Caption Avg. & 53.4 & 75.0 & 81.4 \\
    {\sc eao} ~\cite{shvetsova2022everything} & Caption Avg. & 62.9 & 80.5 & 86.7 \\
    
    \rowcolor[gray]{.95} Ours & Caption Avg. & \textbf{65.1} & \textbf{83.2} & \textbf{87.6} \\
        \bottomrule
    \end{tabular}}
    \vspace{-0.5em}
    \caption{{\small Zero-Shot Text-to-Full Video retrieval on YouCook2. Averaging across captions is used to obtain video-level predictions.}}
    \vspace{-1.5em}
    \label{tab:zs_full_ret}
\end{table}

\noindent{\bf Zero-shot Full Video Retrieval.} In Tab.~\ref{tab:zs_full_ret},  we report results for text-to-full-video retrieval task. Our approach outperforms prior works by $2.2\%, 2.7\%$ on R@1 an R@5 respectively.

\noindent{\bf Retrieval after Fine-tuning.} We also evaluate the retrieval performance of our model after fine-tuning on the downstream datasets as shown in Tab.~\ref{tab:ft_ret}. For a fair comparison, we only report the baselines that use the same training split.
We outperform state-of-the-art consistently on all the metrics and on both MSR-VTT \& YouCook2 datasets as shown in Tab.~\ref{tab:ft_ret}. 

\begin{figure*}[]
\begin{subfigure}[b]{0.45\linewidth}
    \centering
    \includegraphics[width=0.32\linewidth]{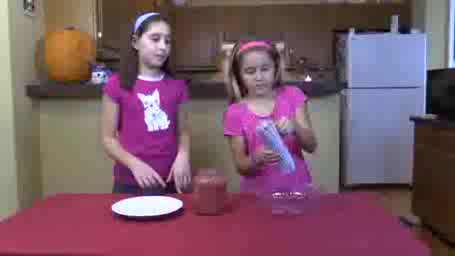}
    \hfill
    \includegraphics[width=0.32\linewidth]{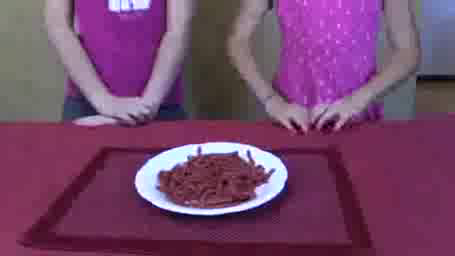}
    \hfill
    \includegraphics[width=0.32\linewidth]{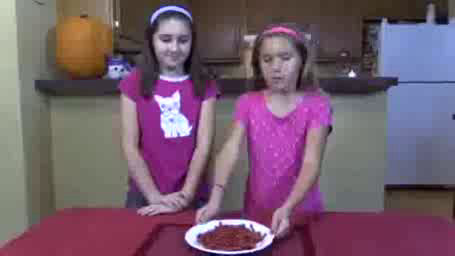}
    \caption{\texttt{UMeazzCYlps}  (Holidays and Traditions)}
\end{subfigure}
\hfill
\begin{subfigure}[b]{0.45\linewidth}
    \centering
    \includegraphics[width=0.32\linewidth]{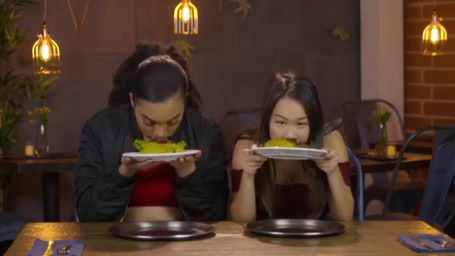}
    \hfill
    \includegraphics[width=0.32\linewidth]{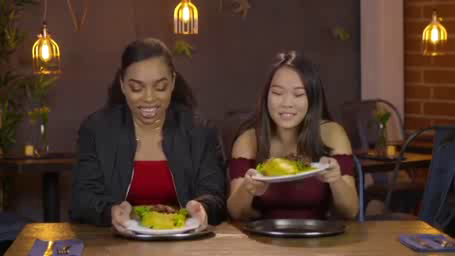}
    \hfill
    \includegraphics[width=0.32\linewidth]{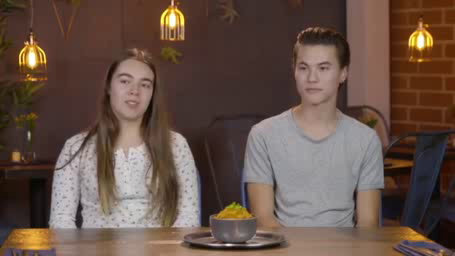}
    \caption{\texttt{fdUgKk-G5Tw} (Food and Entertaining)}
\end{subfigure}
\begin{subfigure}[b]{\linewidth}
    \centering
    \includegraphics[width=\textwidth]{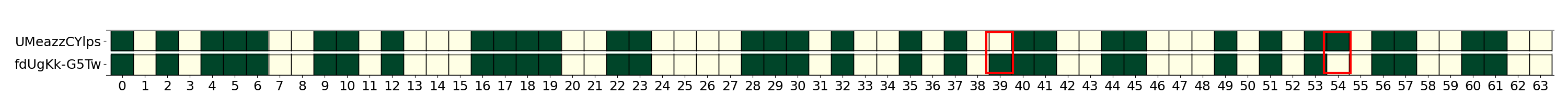}
    \caption{Video Anchor Assignments}
\end{subfigure}
\hfill
\begin{subfigure}[b]{0.45\linewidth}
    \centering
    \includegraphics[width=0.32\linewidth]{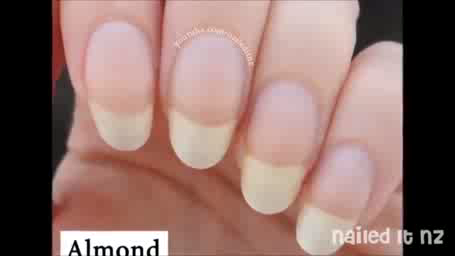}
    \hfill
    \includegraphics[width=0.32\linewidth]{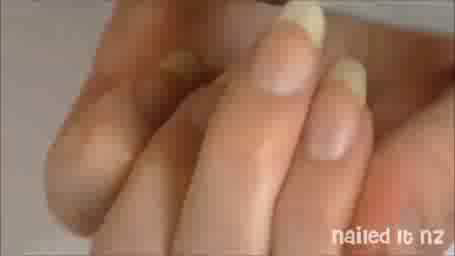}
    \hfill
    \includegraphics[width=0.32\linewidth]{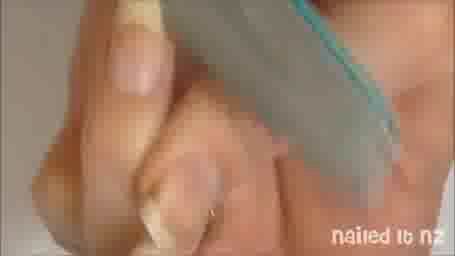}
    \caption{\texttt{A0hURqF7h-c}  (Personal Care and Style)}
\end{subfigure}
\hfill
\begin{subfigure}[b]{0.45\linewidth}
    \centering
    \includegraphics[width=0.32\linewidth]{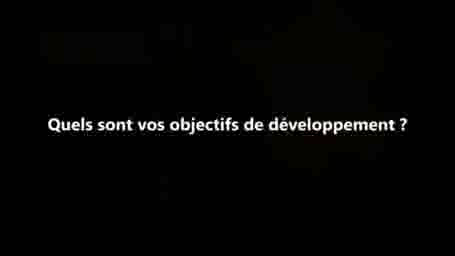}
    \hfill
    \includegraphics[width=0.32\linewidth]{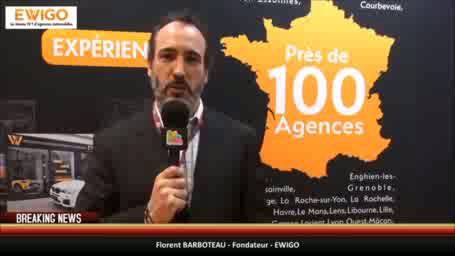}
    \hfill
    \includegraphics[width=0.32\linewidth]{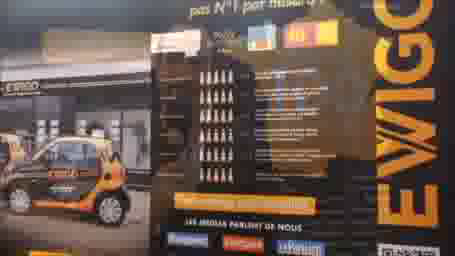}
    \caption{\texttt{\_Y5lyX3KJoE} (Finance and Business)}
\end{subfigure}
\begin{subfigure}[b]{\linewidth}
    \centering
    \includegraphics[width=\textwidth]{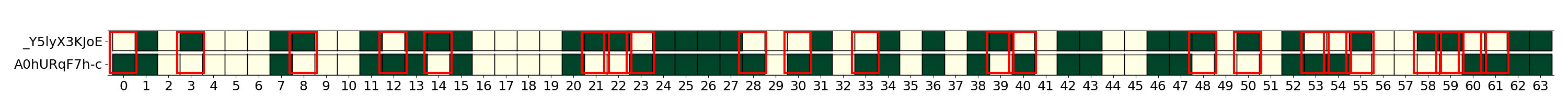}
    \caption{Video anchors Assignments}
\end{subfigure}
\caption{{\small 
Anchor assignments illustrated in this figure demonstrate the visual similarity between video samples from two related categories \ie (a) Holidays and Tradition and (b) Food and Entertaining; and two different categories \ie (d) Personal Care ans Style and (e) Finance and Business  within the HowTo100M dataset. The visual similarity is also reflected in video anchor assignments (c) as most assigned anchors are similar with minor differences in assignments, 
thereby showcasing the flexibility and effectiveness of our approach. (d), (e) samples look very different and therefore the anchor assignments are also very different as shown in (f). Green cell $\rightarrow$ Anchor assigned, Yellow $\rightarrow$ Anchor not assigned. Difference in anchor assignments indicated in \fcolorbox{red}{white}{red}.}}

\label{fig:fg_simcls}
\end{figure*}

\subsection{Ablation Studies} 
\label{sec:ablation}
First we analyze the impact of the proposed components and report results in Tab.~\ref{tab:main_ablation}, followed be effect of number of anchors ($K$) \& ($K'$) in Tab. ~\ref{tab:zs_ret_abl_k_kp}.

\begin{table}[h]
\small
\centering
    \begin{tabular}{l|cc|cc}
        \toprule
    \multirow{2}{*}{Method}  &
    \multicolumn{2}{c}{MSR-VTT} & \multicolumn{2}{c}{YouCook2} \\ 
        \cmidrule{2-5} 
      & {\bf R@5}$\uparrow$ & {\bf R@10}$\uparrow$  & {\bf R@5}$\uparrow$ & {\bf R@10}$\uparrow$ \\ 
    
    \midrule
    Recon. + CL &  23.1 & 32.4 &  37.8 & 48.7   \\
    w/o {\sc cm sspcl} &  22.7  & 31.8   & 36.9  & 48.3   \\
    w/o {\sc sspcl} &  23.8 & 33.3    & 37.9  & 48.8  \\
   
    Modified {\sc sk}
    & 23.4  &  31.3   & 37.9  & 48.3   \\
    
    \rowcolor[gray]{.95} Ours & \textbf{25.1}  & \textbf{34.5} & \textbf{39.4} & \textbf{50.1}  \\
    
    \bottomrule
    \end{tabular}
    \vspace{-0.5em}
     \caption{{\small Ablation studies showing the impact of various components for zero-shot retrieval task. Recon.=Reconstruction Loss, {\sc cm sspcl}=Cross-Modal SSPC Loss, {\sc sk}=Sinkhorn-Knopp,  CL=Contrastive Loss. 
     }}
    \label{tab:main_ablation}
\end{table}

\paragraph{Effect of Proposed Components.} We report the results for this ablation in Tab.~\ref{tab:main_ablation}. In first row, we report results using reconstruction loss and contrastive loss, we notice that using reconstruction loss reduces the performance by $2\%$ on all metrics indicating the effectiveness of the proposed {\sc sspc} loss. In the second row, we report results without our proposed cross-modal {\sc sspc} loss (`w/o {\sc cm sspc}'). We employ this loss to bring the cross-modal representations closer in the joint embedding space to obtain better performance in zero-shot cross-modal tasks. We notice that removing the cross-modal {\sc sspc} loss drastically decreases the zero-shot retrieval performance on both MSR-VTT and YouCook2 datasets, $2.4\%$ and $2.5\%$ drop in R@5 performance respectively. This empirically validates the effectiveness of the cross-modal {\sc sspc} loss in obtaining better cross-modal representations.

\begin{table}[h]
\small
\centering
    \begin{tabular}{c|c|cc|cc}
        \toprule
    \multirow{2}{*}{$K$}  & \multirow{2}{*}{$K'$}  &
    \multicolumn{2}{c}{MSR-VTT} & \multicolumn{2}{c}{YouCook2} \\ 
        \cmidrule{3-6} 
      & & {\bf R@5}$\uparrow$ & {\bf R@10}$\uparrow$  & {\bf R@5}$\uparrow$ & {\bf R@10}$\uparrow$  \\ 
    
    \midrule
    $16$ & $8$  & 23.1 & 32  & 37.1  & 47.5   \\
    $32$ &  $16$  & 23.2 & 32.1  & 36.1  & 47.6  \\
    \midrule
     
    $64$ & $16$ &  23.3 & 31.8  & 36.7  & 47.2  \\
    $64$ & $48$ &  23.7 &32.1  & 36.7 & 47.7   \\
    \rowcolor[gray]{.95} $64$(Ours) & $32$ & \textbf{25.1}  & \textbf{34.5 }& \textbf{39.4} & \textbf{50.1}  \\
    
    \bottomrule
    \end{tabular}
     \caption{{\small Effect of different \# of anchors on zero-shot retrieval. \\ $K$ $\rightarrow$ \# of anchors and $K'$ $\rightarrow$ \# of selected anchors, respectively.  
     }}
    \label{tab:zs_ret_abl_k_kp}
\end{table}

In the third  row, we analyze the effect of the proposed {\sc sspc} loss (`w/o {\sc sspcl}'). We apply the {\sc sspc} loss to retain modality-specific semantic structure from the pre-trained models in the joint embedding space. To investigate its impact, we remove the anchor consistency between the modality-specific and joint embedding spaces. Instead, we enforce the anchor assignments before  \textit{Multi-SK} to be consistent with the \textit{Multi-SK} optimized anchor assignments from the \textit{same} embedding space. We also apply the cross-modal {\sc sspc} loss to only isolate the impact of the proposed {\sc sspc} loss across feature projections. We observe that removing the anchor consistency between the modality-specific and joint embedding spaces reduces the zero-shot retrieval performance by a significant margin ($\sim 1.5\%$ drop in R@5 performance), indicating the importance of {\sc sspc} loss in maintaining modality-specific semantic structure in the joint embedding space for better performance.    

\begin{figure*}[t]
    \centering
    \captionsetup[subfigure]{labelformat=empty}

\begin{subfigure}[b]{0.19\linewidth}
    \caption{``Animated comic scene of guy cutting up food for dinner''}
\end{subfigure}
\hfill
\begin{subfigure}[t]{0.38\linewidth}
    \begin{overpic}[width=0.32\linewidth]{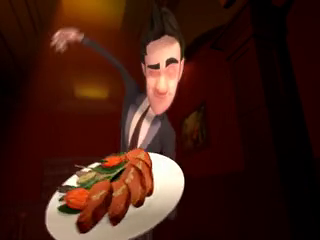}
     \put (27,1) {\colorbox{white}{\textcolor{customgreen}{\scriptsize Match}}}
    \end{overpic}
    \begin{overpic}[width=0.32\linewidth]{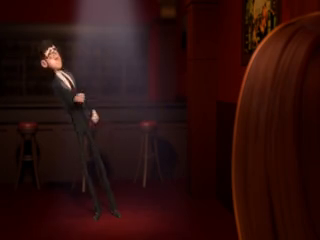}
    \end{overpic}
    \begin{overpic}[width=0.32\linewidth]{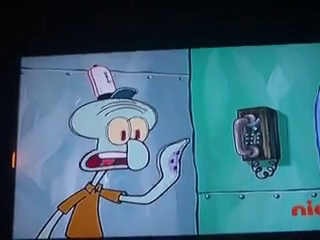}
    \end{overpic}
\end{subfigure}
\hfill
\begin{subfigure}[t]{0.38\linewidth}
    \begin{overpic}[width=0.32\linewidth]{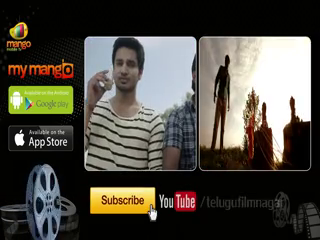}
    \end{overpic}
    \begin{overpic}[width=0.32\linewidth]{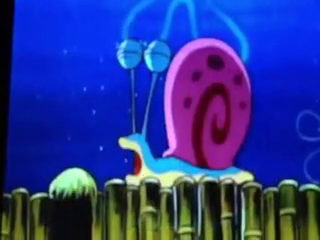}
    \end{overpic}
    \begin{overpic}[width=0.32\linewidth]{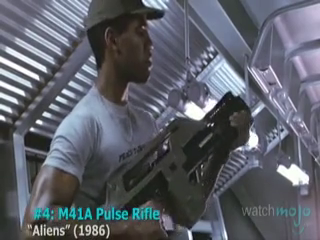}
    \end{overpic}
\end{subfigure}

\begin{subfigure}[b]{0.19\linewidth}
    \caption{``a woman holding a ribbon''}
\end{subfigure}
\hfill
\begin{subfigure}[t]{0.38\linewidth}
    \begin{overpic}[width=0.32\linewidth]{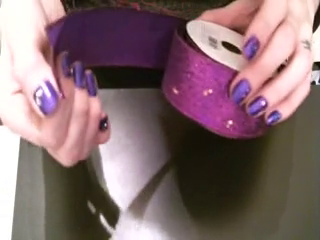}
     \put (27,1) {\colorbox{white}{\textcolor{customgreen}{\scriptsize Match}}}
    \end{overpic}
    \begin{overpic}[width=0.32\linewidth]{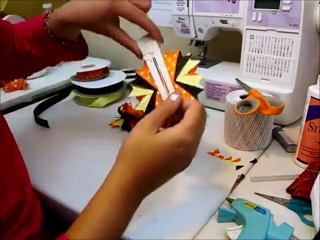}
    \end{overpic}
    \begin{overpic}[width=0.32\linewidth]{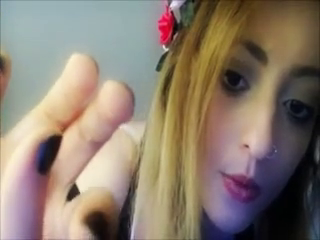}
    \end{overpic}
\end{subfigure}
\hfill
\begin{subfigure}[t]{0.38\linewidth}
    \begin{overpic}[width=0.32\linewidth]{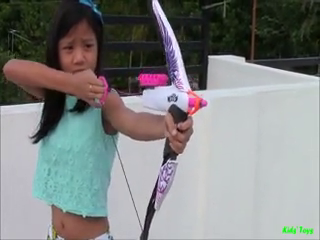}
    \end{overpic}
    \begin{overpic}[width=0.32\linewidth]{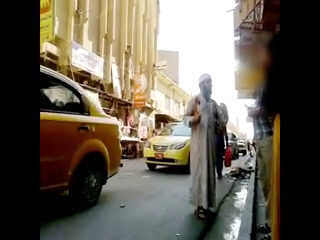}
    \end{overpic}
    \begin{overpic}[width=0.32\linewidth]{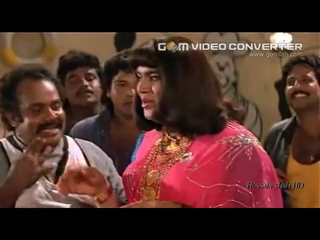}
    \end{overpic}
\end{subfigure}


\begin{subfigure}[b]{0.19\linewidth}
    \caption{``mix ingredients refrigerate''}
\end{subfigure}
\hfill
\begin{subfigure}[t]{0.38\linewidth}
    \begin{overpic}[width=0.32\linewidth]{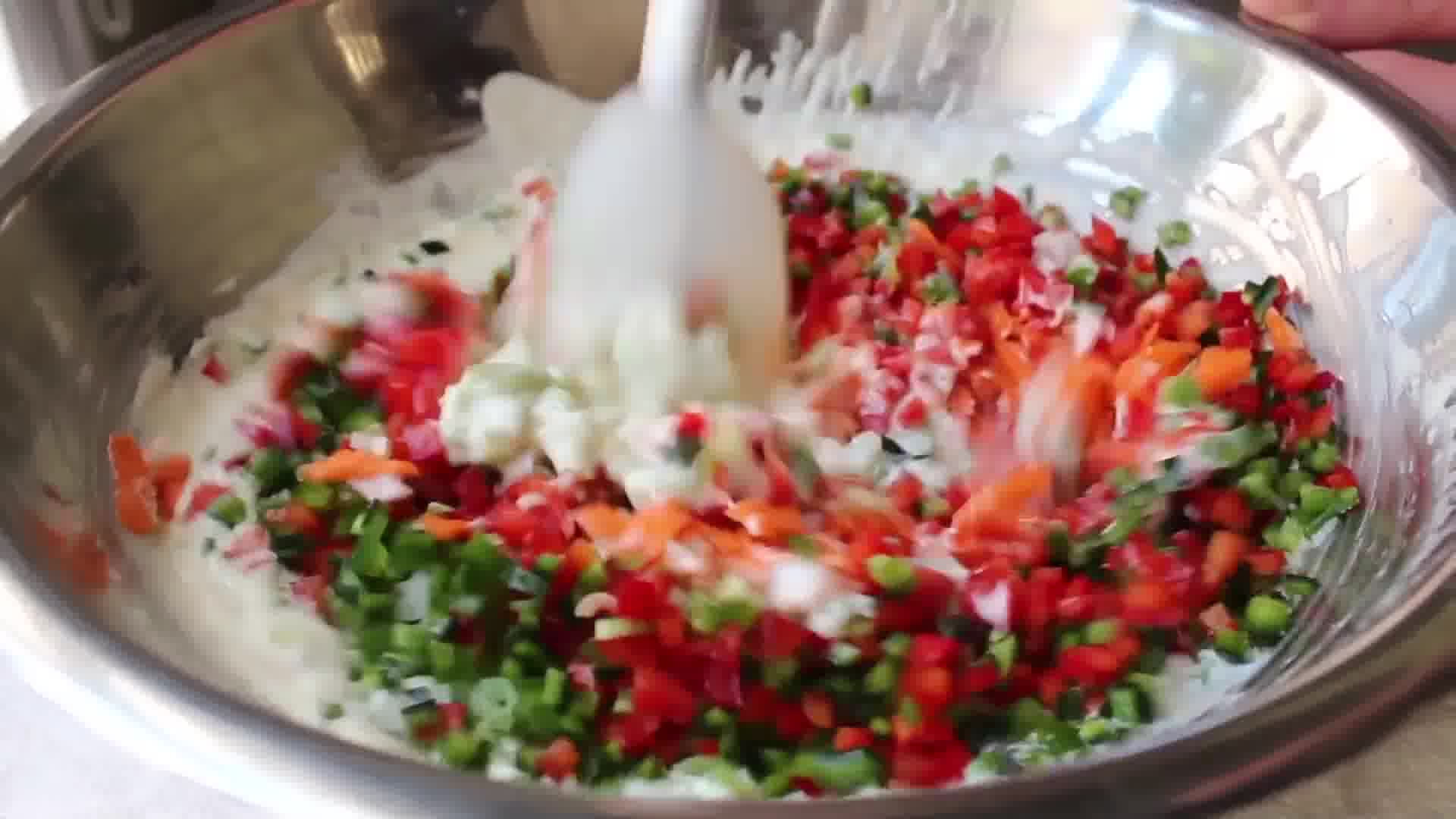}
     \put (27,1) {\colorbox{white}{\textcolor{customgreen}{\scriptsize Match}}}
    \end{overpic}
    \begin{overpic}[width=0.32\linewidth]{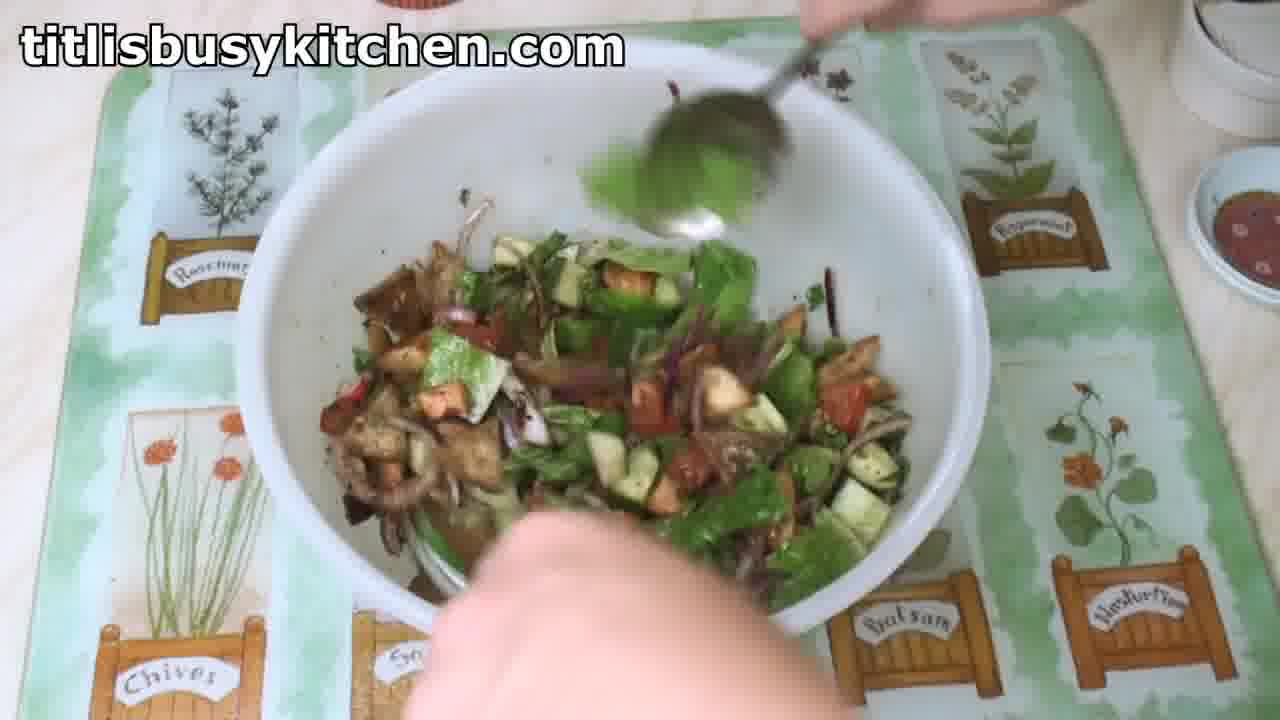}
    \end{overpic}
    \begin{overpic}[width=0.32\linewidth]{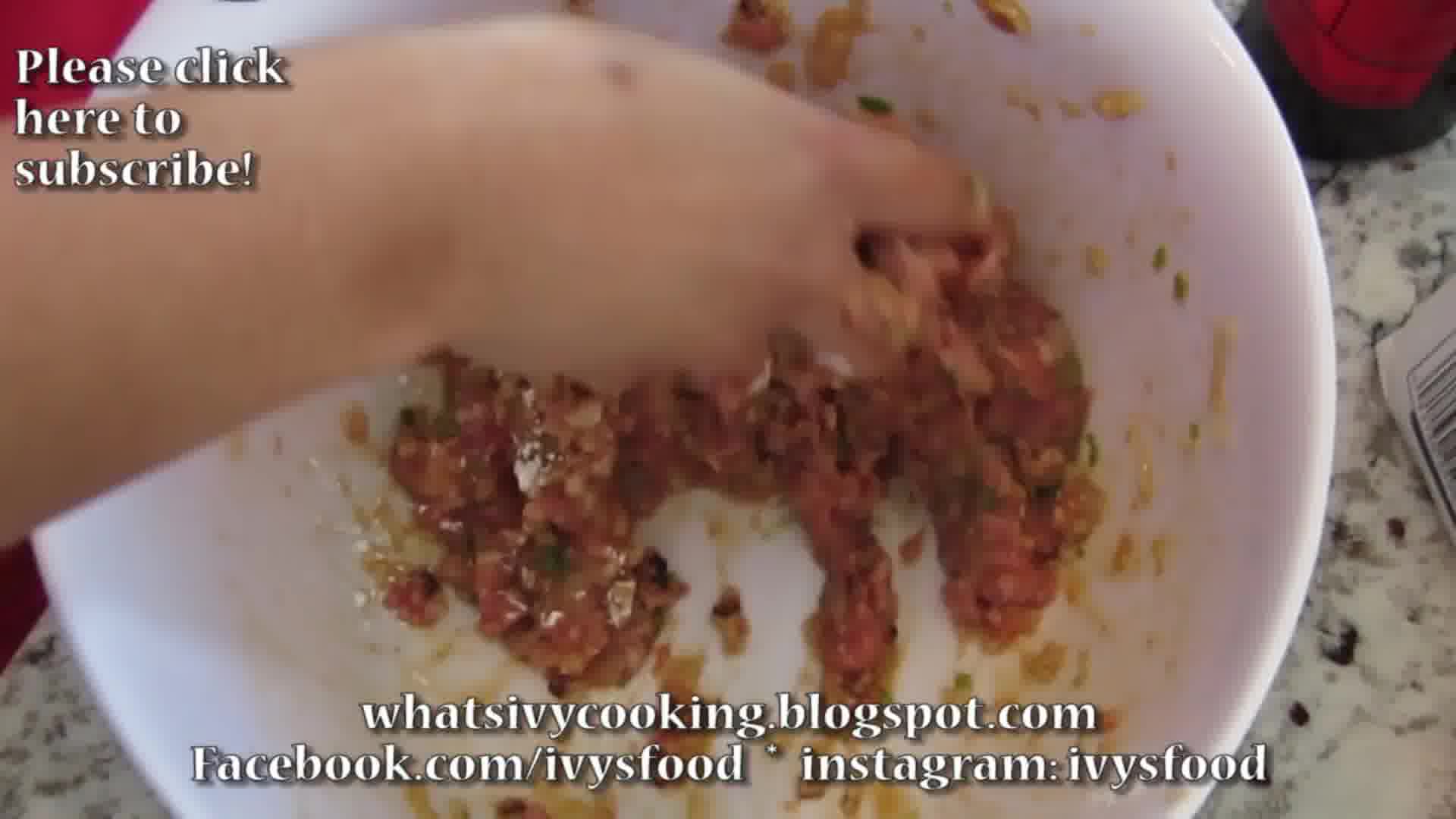}
    \end{overpic}
\end{subfigure}
\hfill
\begin{subfigure}[t]{0.38\linewidth}
    \begin{overpic}[width=0.32\linewidth]{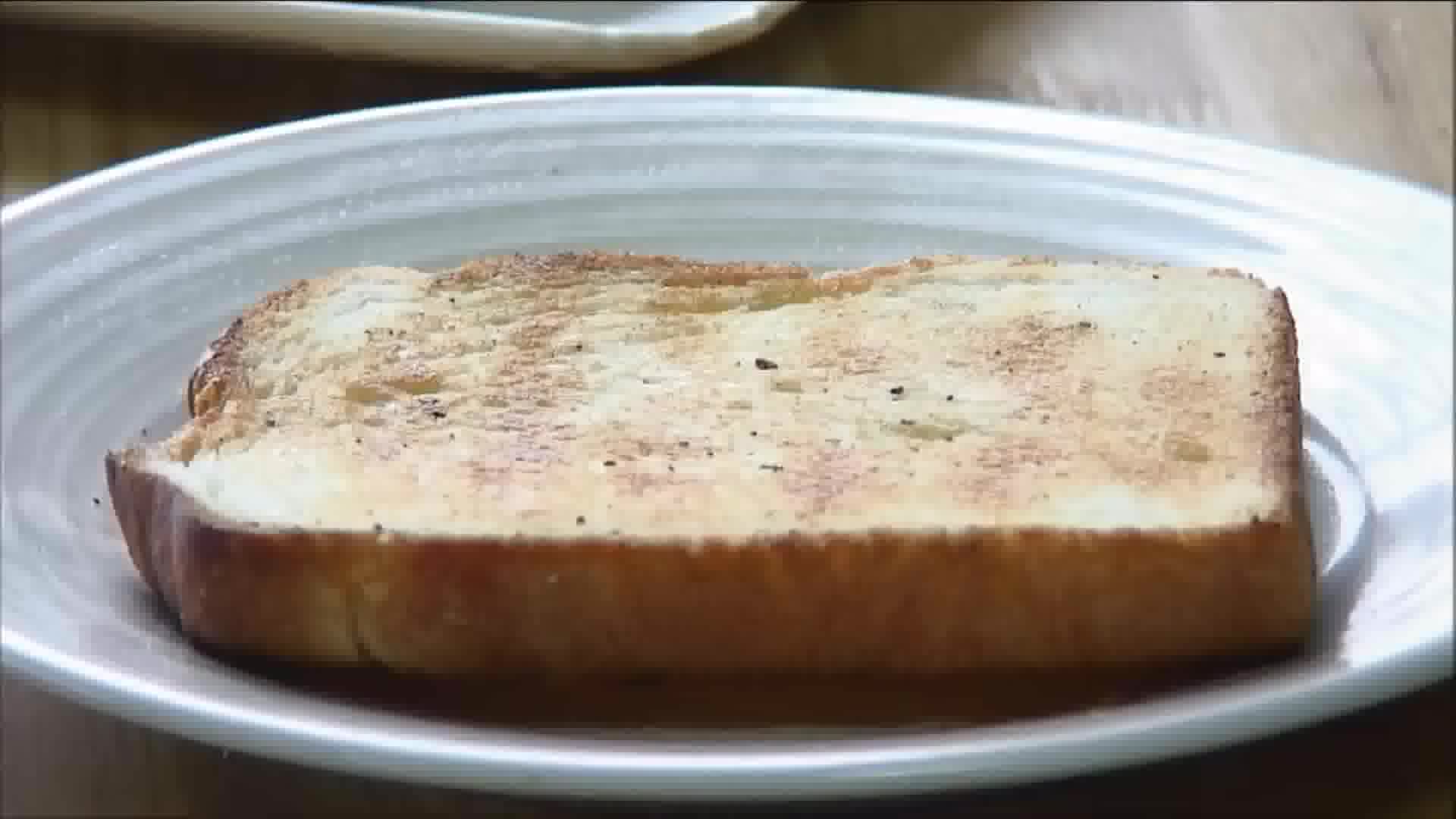}
    \end{overpic}
    \begin{overpic}[width=0.32\linewidth]{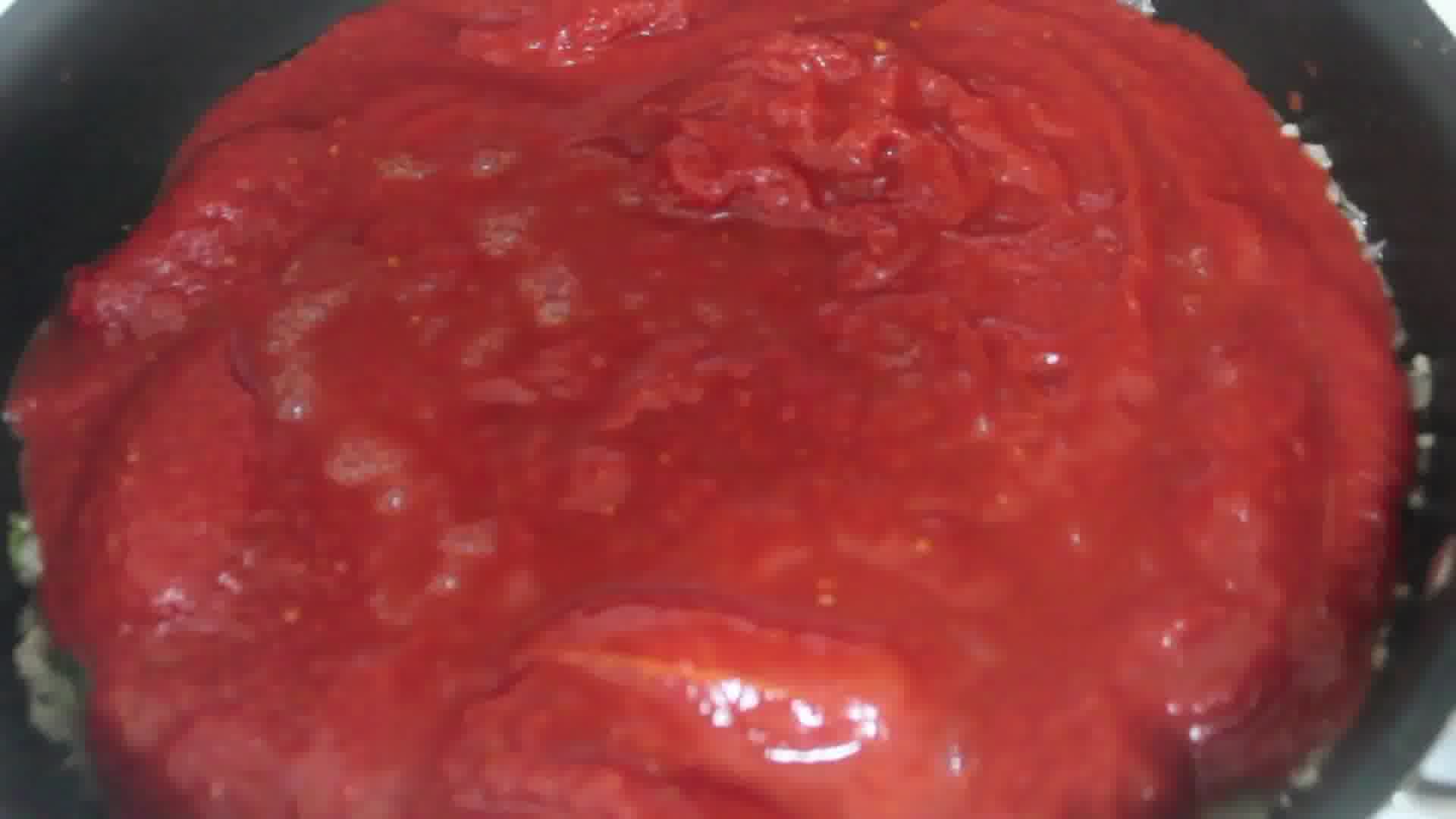}
    \end{overpic}
    \begin{overpic}[width=0.32\linewidth]{images/yc/ffhliBglDhY_176.jpg}
    \end{overpic}
\end{subfigure}

\begin{subfigure}[b]{0.19\linewidth}
    \caption{``add mutton pan''}
\end{subfigure}
\hfill
\begin{subfigure}[t]{0.38\linewidth}
    \begin{overpic}[width=0.32\linewidth]{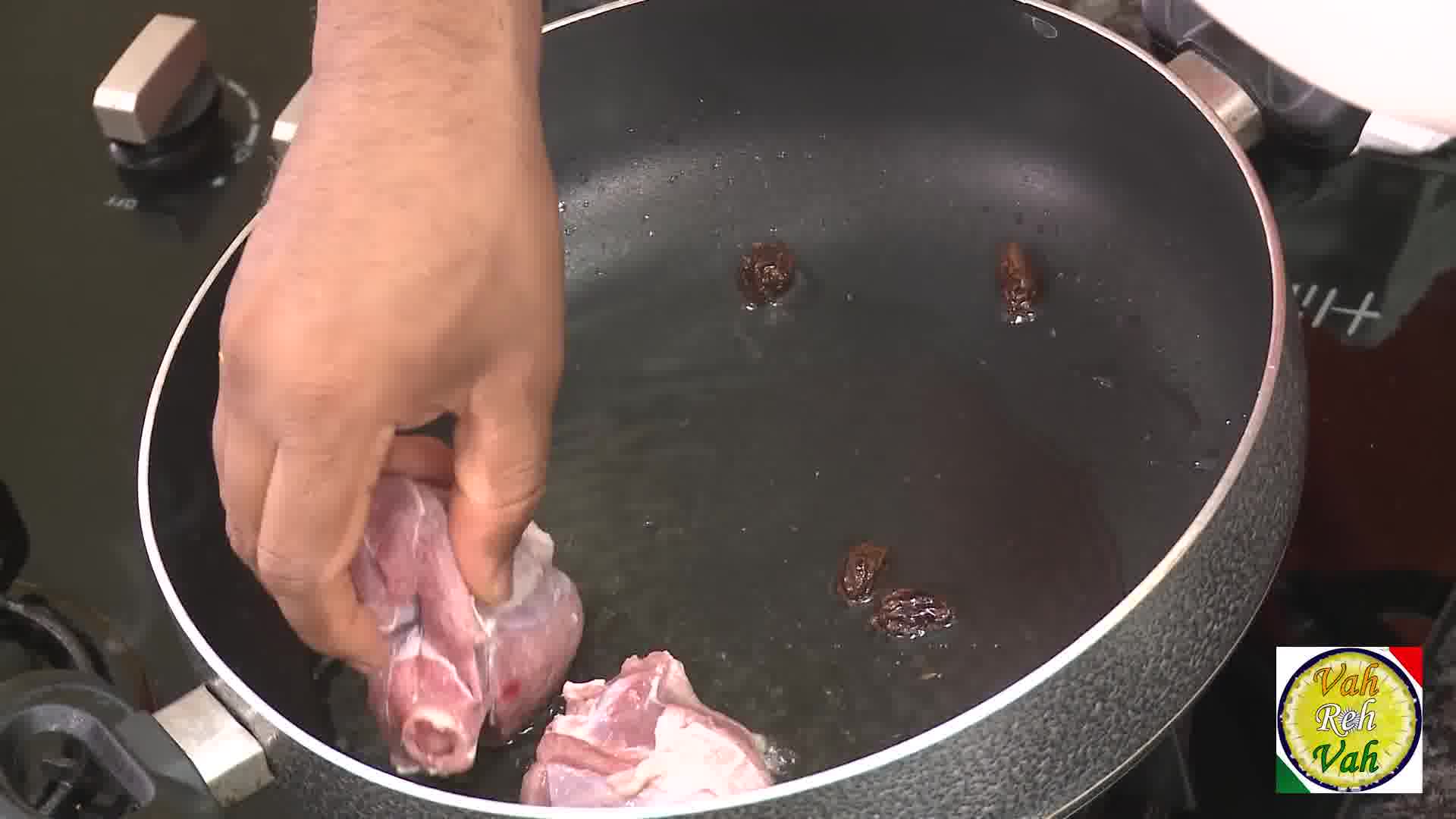}
     \put (27,1) {\colorbox{white}{\textcolor{customgreen}{\scriptsize Match}}}
    \end{overpic}
    \begin{overpic}[width=0.32\linewidth]{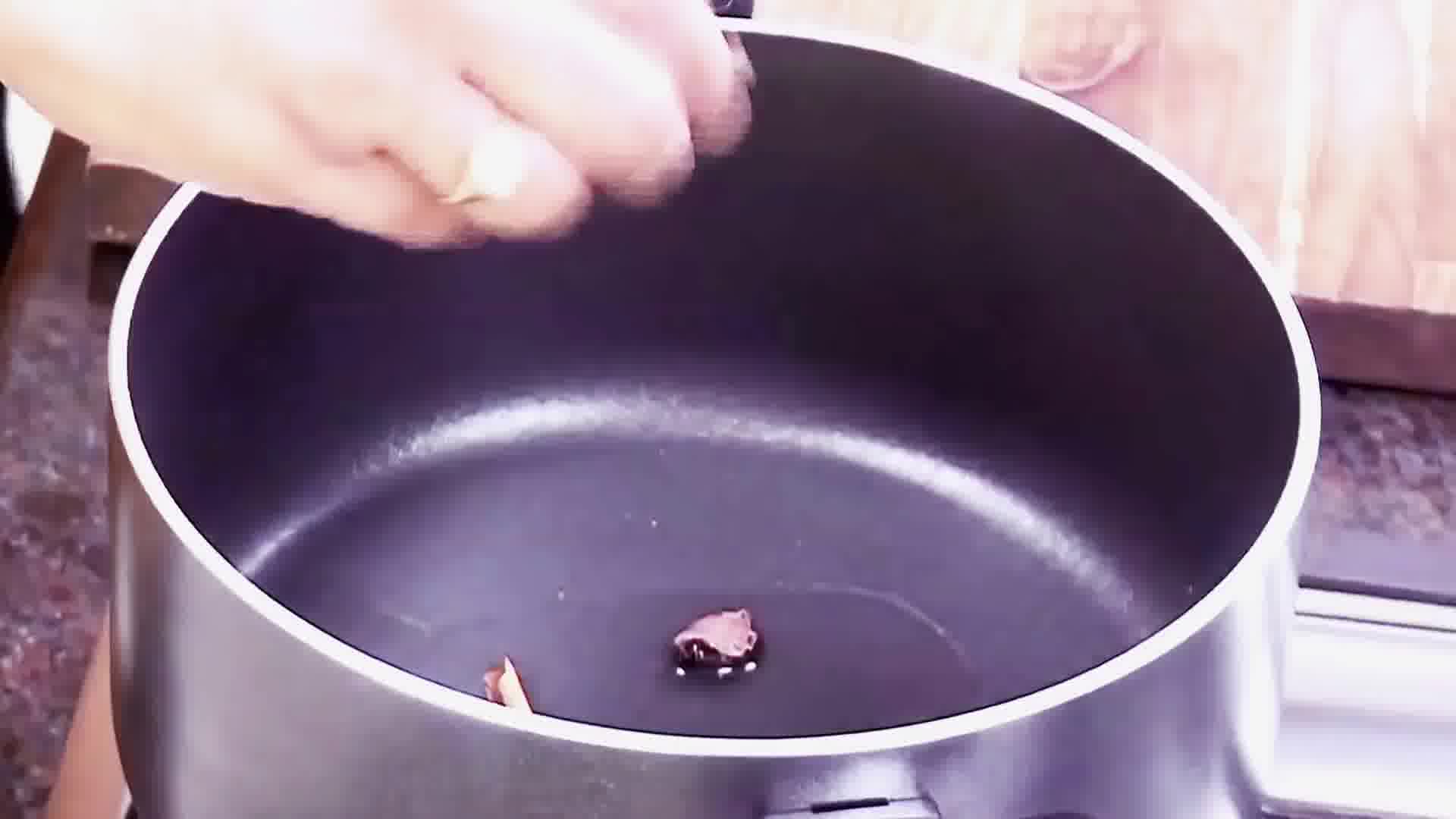}
    \end{overpic}
    \begin{overpic}[width=0.32\linewidth]{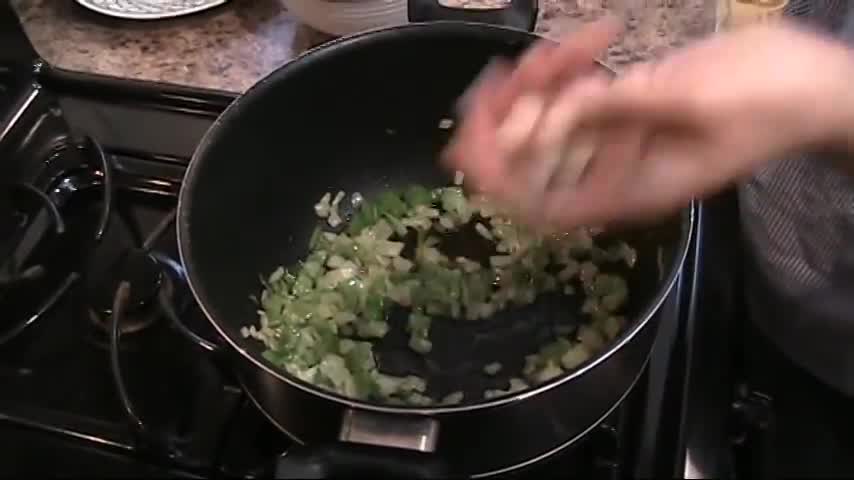}
    \end{overpic}
    \caption{Ours}
\end{subfigure}
\hfill
\begin{subfigure}[t]{0.38\linewidth}
    \begin{overpic}[width=0.32\linewidth]{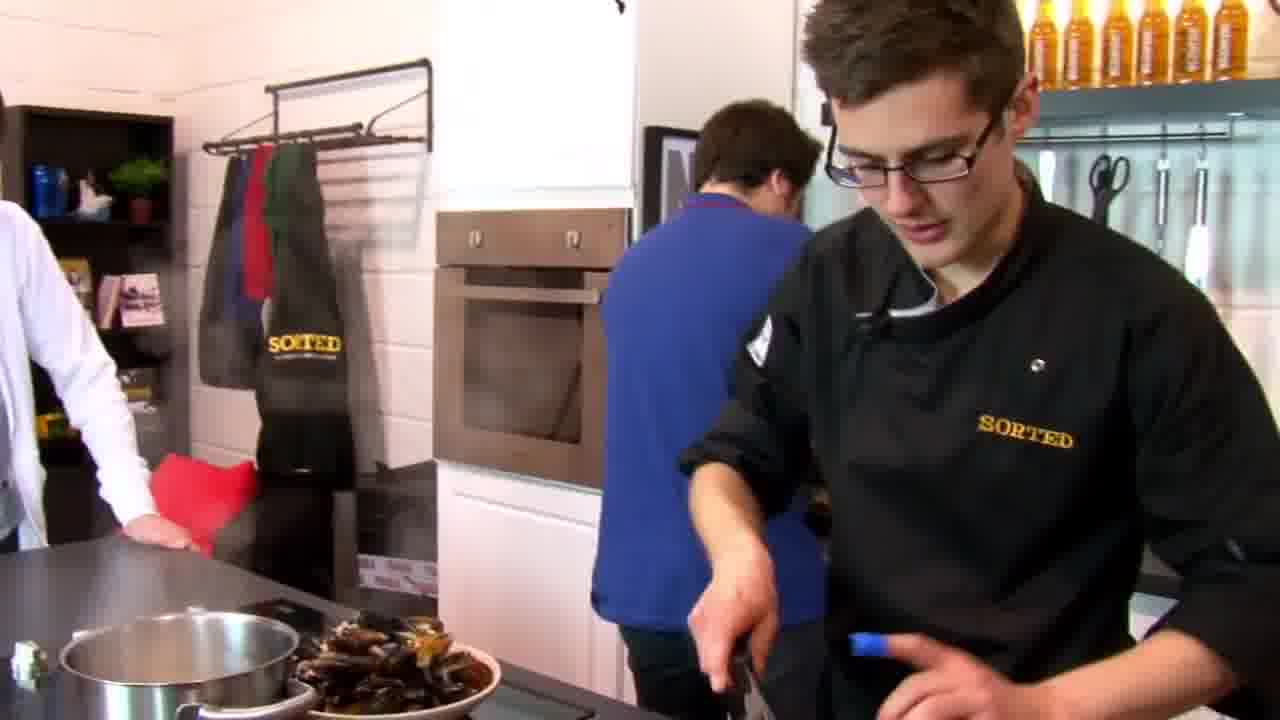}
    \end{overpic}
    \begin{overpic}[width=0.32\linewidth]{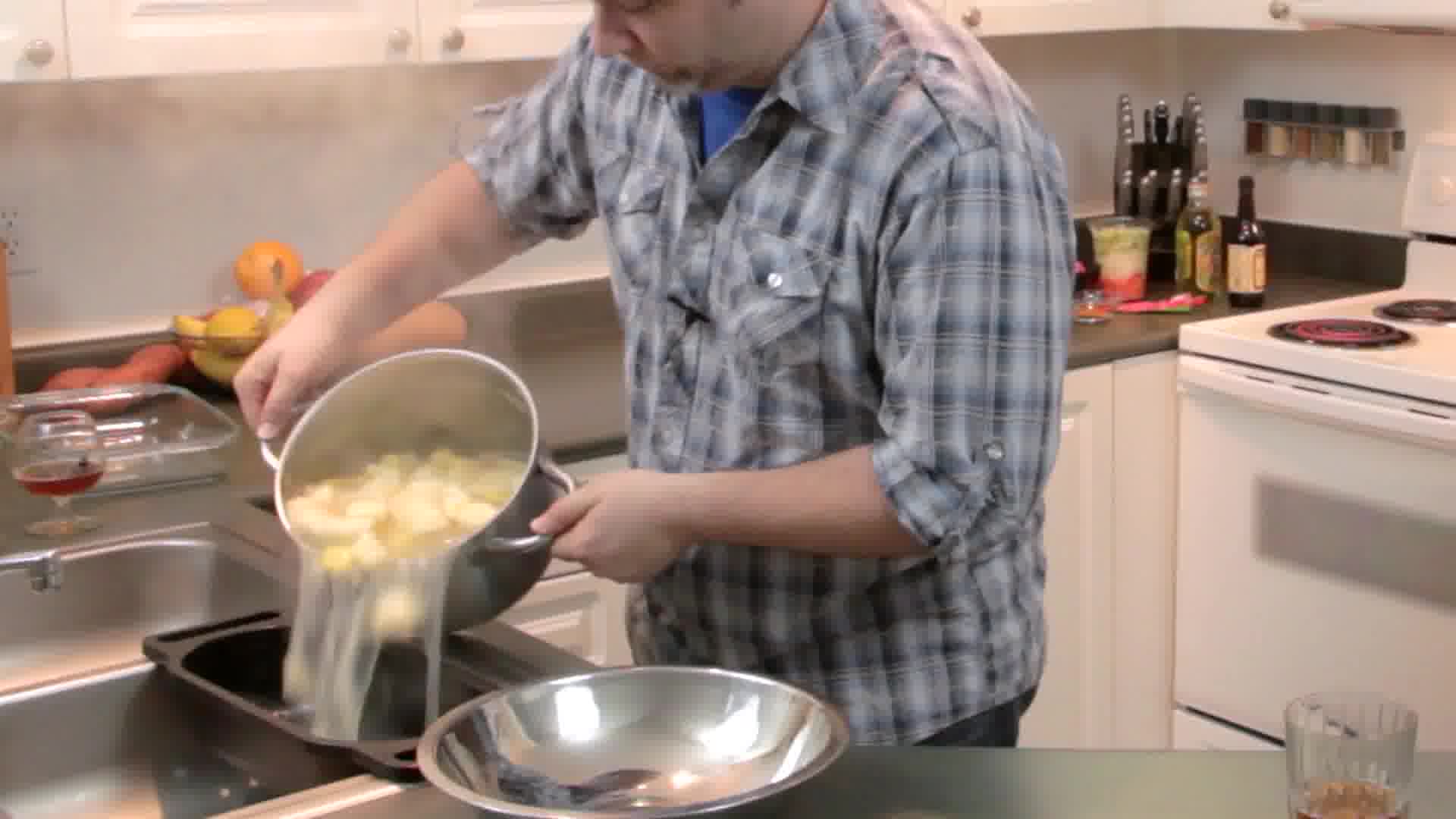}
    \end{overpic}
    \begin{overpic}[width=0.32\linewidth]{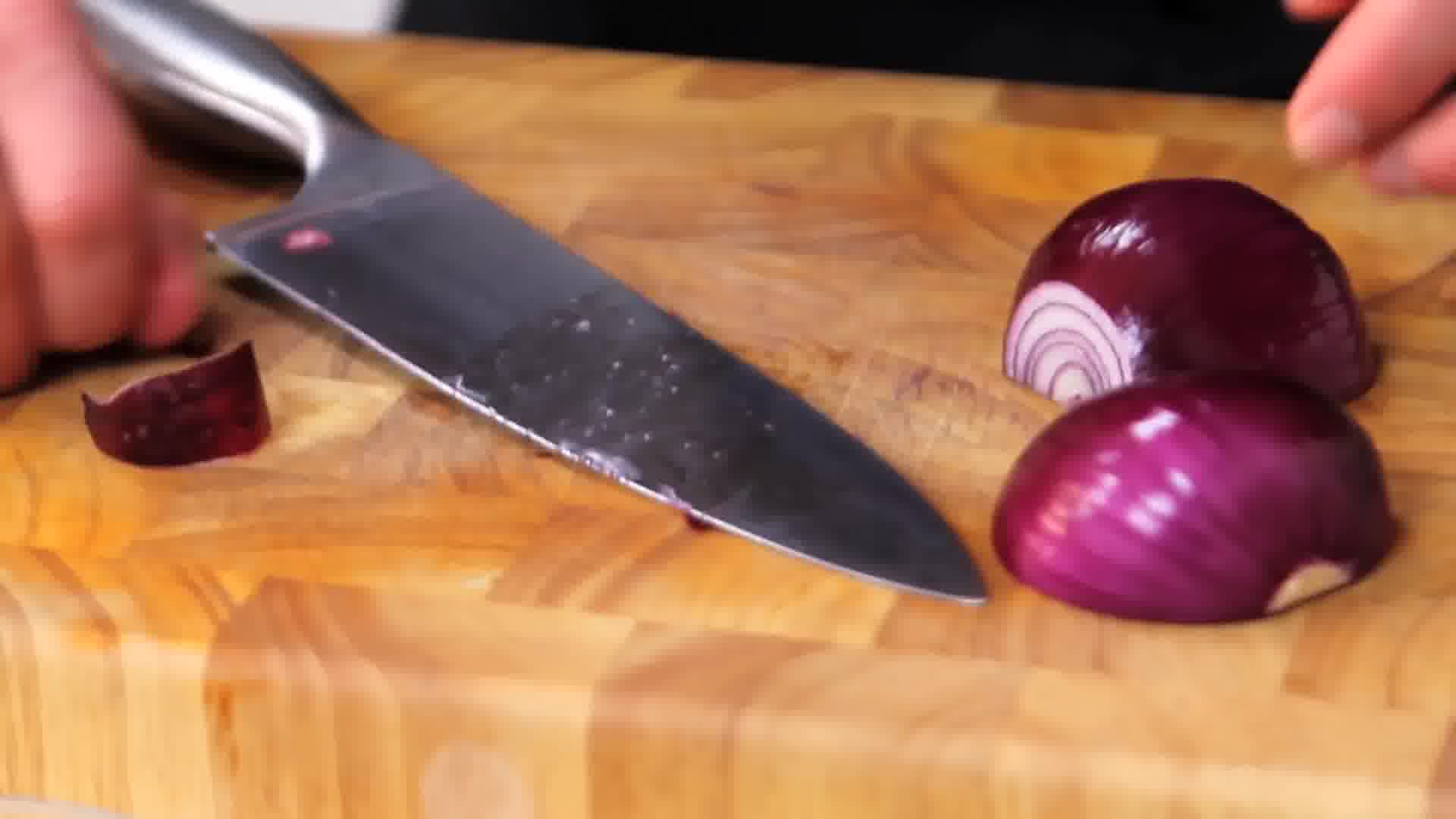}
    \end{overpic}
    \caption{EAO~\cite{shvetsova2022everything}}
\end{subfigure}
\caption{Examples of Zero-Shot Text-to-Video Retrieval on \textbf{MSR-VTT} and \textbf{YouCook2} datasets. Each row consists of Textual Query (left), and top-$3$ retrieved videos for our method (center) and the state-of-the-art method EAO~\cite{shvetsova2022everything}(right). \textit{Match} indicates correct video for the query.}
\label{fig:zs_comp}
\end{figure*}

Finally, in the fourth row of Tab.~\ref{tab:main_ablation}, we analyse the effectiveness of the proposed Multi-Assignment Sinkhorn-Knopp 
algorithm. To this end, we modify the vanilla Sinkhorn-Knopp~\cite{cuturi2013sinkhorn} to obtain multiple anchor assignments per sample (refer to Sec. 5 of the Supplementary for details). 
We notice that the modified Sinkhorn-Knopp does not perform well with a drop in performance ($\sim 1.6\%$ drop in R@5) on both MSR-VTT and YouCook2 datasets on all the metrics indicating the significance of our proposed {\sc multi-sk} algorithm.  


\vspace{-0.5cm}
\paragraph{Effect of Number of Anchors.} To analyze the effect of the number of anchors, we conduct experiments with different number of anchors, $K$, and different number of selected anchors, $K'$. We report the results in Tab.~\ref{tab:zs_ret_abl_k_kp}. We conduct experiments on both MSR-VTT and YouCook2 datasets. From the top half of Tab.~\ref{tab:zs_ret_abl_k_kp}, we observe that the performance of our proposed method improves with the increasing number of anchors. This is to be expected since a higher number of anchors have a higher representation learning capacity. We also observe that the method performs reasonably well even with a very small number of anchors showing the general effectiveness of our proposed solution. We notice that the performance of the proposed method improves as we select more anchors as shown in bottom half of Tab.~\ref{tab:zs_ret_abl_k_kp}. However, selecting a very large number of anchors ($48$ out of $64$ for this experiment) adds more constraints leading to poor performance.

\subsection{Qualitative Analysis} Here, first we present a fine-grained visual analysis of the learned anchors, followed by qualitative retrieval results.

\par We show fine-grained analysis of the learned anchors in Fig.~\ref{fig:fg_simcls}. For the purpose of this analysis, we visualize the anchor assignments as binary assignments. However, during training we use soft anchor assignments. In Fig.~\ref{fig:fg_simcls}(c), we compare the anchor assignments for samples from \emph{similar categories}. It can be seen that the videos are visually similar even though they belong to different categories and the anchor assignments for these two examples are able to capture the sample similarity. In Fig.~\ref{fig:fg_simcls}(f), we compare the anchor assignments for videos from \emph{different categories} and it can be seen that the anchor assignments are very different as expected. This further validates our claim that our proposed method can assign semantically meaningful anchors without any explicit supervision. Further, we show qualitative retrieval comparison with EAO~\cite{shvetsova2022everything} in Fig.~\ref{fig:zs_comp}. We present more qualitative analysis in Supplementary Sec. 4.

\section{Conclusion}
\par We proposed a novel approach that preserves the modality-specific semantic relationship between samples in the joint multi-modal embedding space. To this end, we propose a flexible sample relationship modeling approach by assigning multiple anchors to each sample, which captures both shared and unique aspects of samples. To obtain these assignments, we develop a novel \textit{Multi-Assignment Sinkhorn-Knopp} (Multi-SK) algorithm, and also utilize the proposed anchor consistency loss to learn these anchors. Our qualitative results demonstrate that our learnt anchors correspond to meaningful semantic concepts.  Our extensive experimentation demonstrates that the proposed approach improves generalizability by outperforming  state-of-the-art methods on \textit{both} in- and out-of-domain datasets. We also show that our method achieves state-of-the-art performance on multiple zero-shot tasks, and also outperforms when fine-tuned on downstream datasets.


\section*{Acknowledgements}  
Research was sponsored by the Army Research Office and was accomplished under Grant Number W911NF-19-1-0356. The views and conclusions contained in this document are those of the authors and should not be interpreted as representing the official policies, either expressed or implied, of the Army Research Office or the U.S. Government. The U.S. Government is authorized to reproduce and distribute reprints for Government purposes notwithstanding any copyright notation herein. Nina Shvetsova is supported by German Federal Ministry of Education and Research (BMBF) project STCL - 01IS22067.
{\small
\bibliographystyle{iccv}
\bibliography{iccv}
}

\end{document}